\documentclass[9pt,sigconf]{acmart} %

\usepackage{booktabs} 
\usepackage{CJK}
\usepackage{algorithmicx}
\usepackage[ruled,vlined,linesnumbered]{algorithm2e}
\usepackage{hhline}
\usepackage{amsmath,mathtools}
\usepackage{amsfonts}
\usepackage{array}
\usepackage{mathrsfs}
\usepackage{hyperref}
\usepackage{graphicx}
\usepackage{subcaption}
\usepackage{enumitem,color}

\acmYear{2022}\copyrightyear{2022}
\acmConference[CoNEXT '22]{The 18th International Conference on emerging Networking EXperiments and Technologies}{December 6--9, 2022}{Roma, Italy}
\acmBooktitle{The 18th International Conference on emerging Networking EXperiments and Technologies (CoNEXT '22), December 6--9, 2022, Roma, Italy}
\acmPrice{15.00}
\acmDOI{10.1145/3555050.3569115}
\acmISBN{978-1-4503-9508-3/22/12}

\begin{document}
\title[Atlas]{Atlas: Automate Online Service Configuration in Network Slicing}


\author{Qiang Liu}
\affiliation{%
  \institution{University of Nebraska-Lincoln}
}
\email{qiang.liu@unl.edu}

\author{Nakjung Choi}
\affiliation{%
  \institution{Nokia Bell Labs}
}
\email{nakjung.choi@nokia-bell-labs.com}

\author{Tao Han}
\affiliation{%
  \institution{New Jersey Institute of Technology}
}
\email{tao.han@njit.edu}


\renewcommand{\shortauthors}{Qiang Liu, Nakjung Choi and Tao Han}

\begin{abstract}
Network slicing achieves cost-efficient slice customization to support heterogeneous applications and services.
Configuring cross-domain resources to end-to-end slices based on service-level agreements, however, is challenging, due to the complicated underlying correlations and the simulation-to-reality discrepancy between simulators and real networks.
In this paper, we propose Atlas, an online network slicing system, which automates the service configuration of slices via safe and sample-efficient learn-to-configure approaches in three interrelated stages.
First, we design a learning-based simulator to reduce the sim-to-real discrepancy, which is accomplished by a new parameter searching method based on Bayesian optimization.
Second, we offline train the policy in the augmented simulator via a novel offline algorithm with a Bayesian neural network and parallel Thompson sampling.  
Third, we online learn the policy in real networks with a novel online algorithm with safe exploration and Gaussian process regression. 
We implement Atlas on an end-to-end network prototype based on OpenAirInterface RAN, OpenDayLight SDN transport, OpenAir-CN core network, and Docker-based edge server.
Experimental results show that, compared to state-of-the-art solutions, Atlas achieves 63.9\% and 85.7\% regret reduction on resource usage and slice quality of experience during the online learning stage, respectively.
\end{abstract}



\maketitle

\vspace{-0.1in} \section{Introduction}

Network slicing is one of the key building blocks in 5G and Beyond~\cite{foukas2017network} to provide guaranteed networking performances for concurrently supporting various network services and applications, e.g., mobile augmented reality~\cite{liu2018dare}, autonomous driving~\cite{wang2018networking}, and federated learning~\cite{tran2019federated}.
It enables network operators to cost-efficiently create virtual networks (\emph{aka.} network slices) with performance and functional isolation~\cite{foukas2017orion} based on the common physical infrastructure.
Each network slice can be highly customized according to the needs of individual slice tenants~\cite{GSMA}, e.g., throughput per slice user, delay, and quality of service (QoS).
As emerging services are increasingly focusing on end-to-end performances, e.g., round-trip latency, end-to-end slicing is more than ever needed, which consists of the subnet instance~\cite{badmus2019network} in radio access networks (RAN), transport networks (TN), core networks (CN), and edge networks (EN).

\begin{figure}
    \centering
    \includegraphics[width=3.0in, height=1.1in]{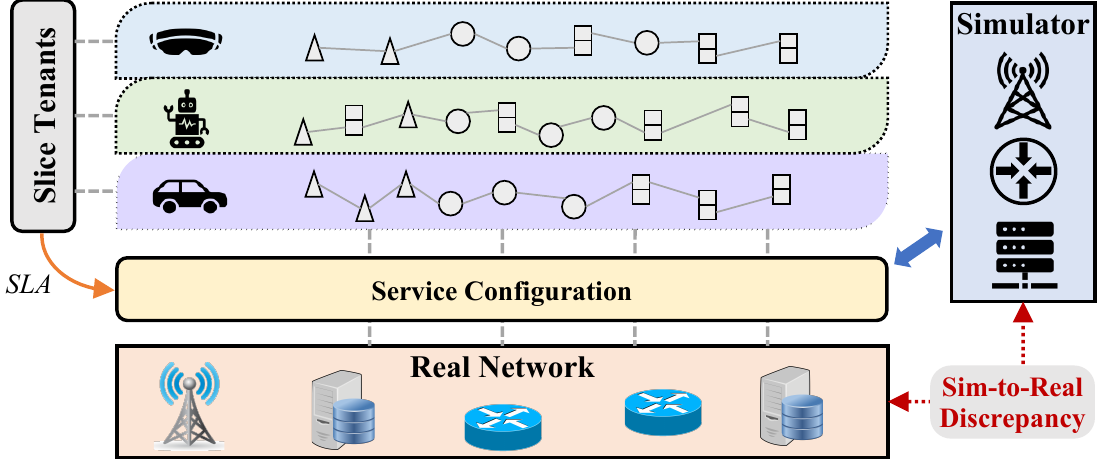}
    \caption{\small Illustration of online network slicing}
    \label{fig:illustration}
\end{figure}

To initialize a network service, slice tenants make service-level agreements (SLAs) with the network operator to specify the performance requirement of its users.
To maintain the performance of individual slices, as shown in Fig.~\ref{fig:illustration}, the network operator aims to derive a policy\footnote{We refer a policy as the mapping from network states, e.g., spatiotemporal user traffic, to service configuration actions, e.g., radio bandwidth.} to dynamically configure the slice's cross-domain resources and settings, e.g., radio and backhaul bandwidth, under varying network dynamics.
As heterogeneous slices are with highly diverse needs with performance metrics, e.g., reliability in vehicle-to-everything and frame-per-second for video streaming~\cite{zhang2020onrl}, it is impractical to comprehensively examine individual slices before their deployments.
Hence, automated network slicing is indispensable to automatically learn the actual needs of slices according to their service-level agreements (SLAs) and intelligently adjust service configuration of slices by exploiting state-of-the-art machine learning (ML) techniques~\cite{velasco2021end, pang2020survey}.



It is, however, challenging to obtain the automated service configuration policy in end-to-end network slicing, even if recent advances in deep neural networks (DNNs) showed promising capability in complex function approximations~\cite{shi2021adapting, mao2019learning, liu2020edgeslice}.
On the one hand, it is unsafe to learn the configuration policy via online interaction with real networks.
The exploration during policy learning may decrease service performances occasionally~\cite{liu2021constraint}, which results in the violation of slice SLA. 
On the other hand, it is insufficient to learn the configuration policy via offline interaction with network simulators.
The simulation-to-reality (sim-to-real) discrepancy between simulators and real networks could compromise the online performance achieved by offline policies~\cite{liu2021onslicing}, even if they seem to perform well in simulators.
The sim-to-real discrepancy exists in different systems, e.g., robotics~\cite{tan2018sim} and networks~\cite{mao2019learning}, which is found to be independent to the type of simulators~\cite{shi2021adapting}.

In this paper, we propose Atlas, an integrated offline-online network slicing system, to automate the service configuration of slices and online learn the policy safely.
Atlas is accomplished with the following novel designs.

\textbf{Learning-based simulator.}
We design a learning-based simulator, whose simulation parameters (e.g., transmission power and operating spectrum of base stations) can be slightly adjusted to match that of real networks for reducing the sim-to-real discrepancy.
Simulators are built with domain knowledge to mimic real networks in multiple aspects, e.g., parameters and protocols.
We observe that the sim-to-real discrepancy is partially attributed to the inaccurate simulation parameter settings, e.g., radio channel models in simulators are usually simplified and abstracted~\cite{tuomaala2005effective}.
Due to the high-dim search space and non-trivial execution time of simulators, conventional searching methods (e.g., exhaustive and grid search) fails under given time periods.
We propose a new method based on Bayesian optimization to search for the optimal simulation parameters, which balances the reduction of sim-to-real discrepancy and the explainability of simulation parameters.
The method is composed of Bayesian neural networks (BNN) as the approximation function, Thompson sampling for trading off exploration and exploitation, and parallel queries for accelerating the convergence with multiprocessing techniques.
The obtained simulation parameters will be set in the simulator, which serves as the offline environment for offline policy training.

\textbf{Offline policy training.}
We propose a novel offline training algorithm to automatically learn to configure while assuring the slice SLA by interacting with the augmented simulator.
Due to the lack of prior models for heterogeneous slices, we exploit BNN to approximate the complex correlation between resource configurations and slice performances.
To assure the slice SLA, we design an adaptive penalization method to incorporate the weighted constraint into the objective with a dynamic multiplier.
As the interdependencies between consecutive configurations are weakened by large intervals~\cite{shi2021adapting, marquez2018should}, e.g., 1 hour, we adopt Bayesian optimization with parallel Thompson sampling to search the optimal offline policy that is utilized to accelerate online learning.

\textbf{Online policy learning.}
We propose a novel online learning algorithm to safely and sample-efficiently learn the optimal online policy and resolve the sim-to-real discrepancy within real networks.
First, we adopt the sample-efficient Gaussian process as the approximation function, and use it to learn the sim-to-real discrepancy only, which is simpler than the whole correlation between resource configurations and slice performances.
Second, we design a conservative acquisition function that achieves safe exploration with the guarantee of Bayesian regret bound.
Third, we exploit the offline simulator to augment online transitions and update the multiplier for accelerating the online learning progress.

\textbf{Contributions.}
To the best of our knowledge, Atlas is the first online network slicing system that automatically learn to configure while assuring the slice SLA.
The specific contributions of Atlas are summarized as follows:
\begin{itemize}[leftmargin=*]
    \item We design a new parameter searching method (Sec.~\ref{sec:learning_based_simulator}) to automatically search the simulation parameters for offline simulators to reduce the sim-to-real discrepancy.
    \item We design a novel offline training algorithm (Sec.~\ref{sec:offline_training}) to automate the service configuration of slices in the augmented simulator. 
    \item We design a novel online learning algorithm (Sec.~\ref{sec:online_learning}) to safely and sample-efficiently learn the online policy and resolve the sim-to-real discrepancy within real networks.
    \item We implement Atlas on an end-to-end network prototype (Sec.~\ref{sec:implementation}) and conduct extensive experiments to evaluate Atlas in terms of performance assurance and sample-efficiency (Sec.~\ref{sec:evaluation}).
\end{itemize}














\begin{figure*}[!t] 
\captionsetup{justification=centering}
  \begin{minipage}[t]{0.245\textwidth}
    \centering
    \includegraphics[width=1.75in, height=1.2in]{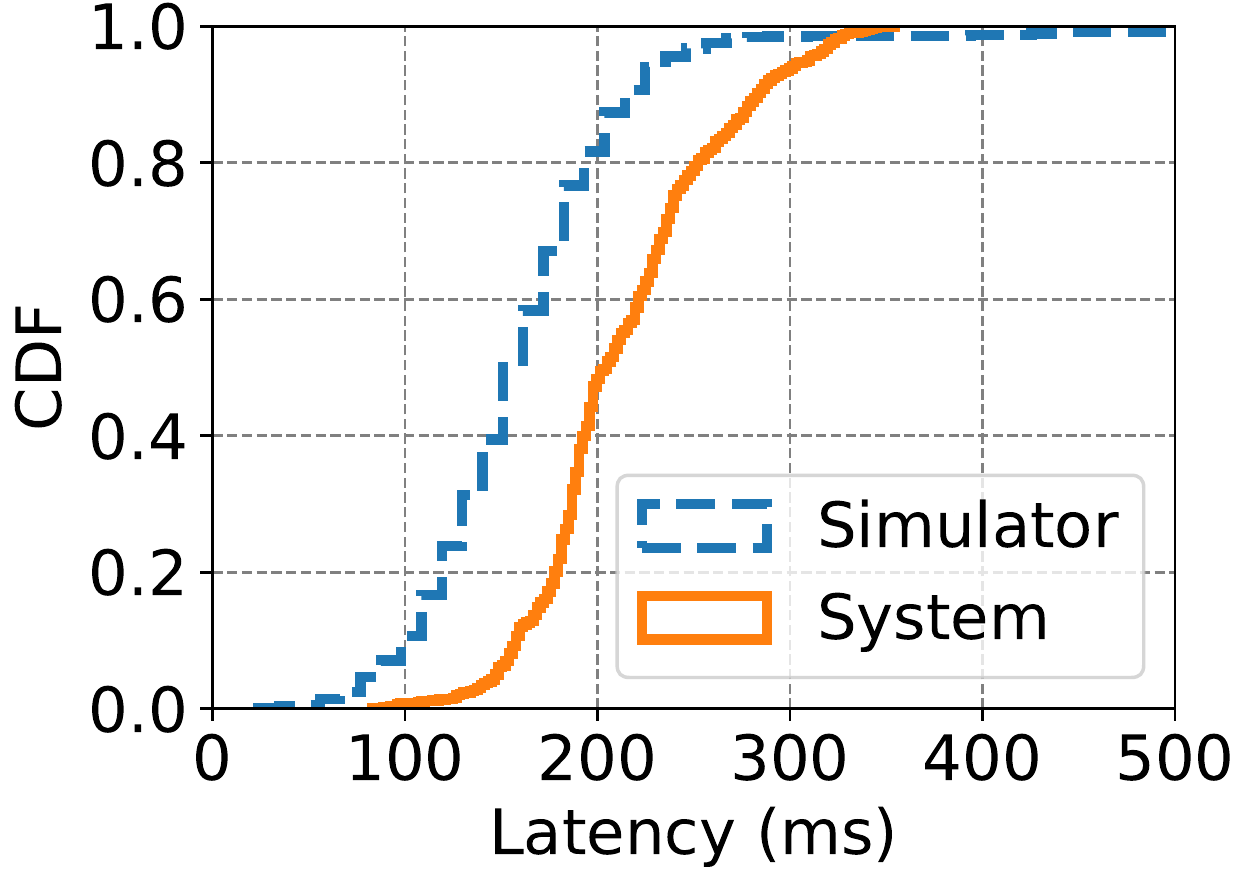}
    \caption{\small End-to-end latency under one slice user}
    \label{fig:result_motivation_sim_to_real_mar_traffic_1}
  \end{minipage}
  \begin{minipage}[t]{0.245\textwidth}
    \centering
    \includegraphics[width=1.75in, height=1.2in]{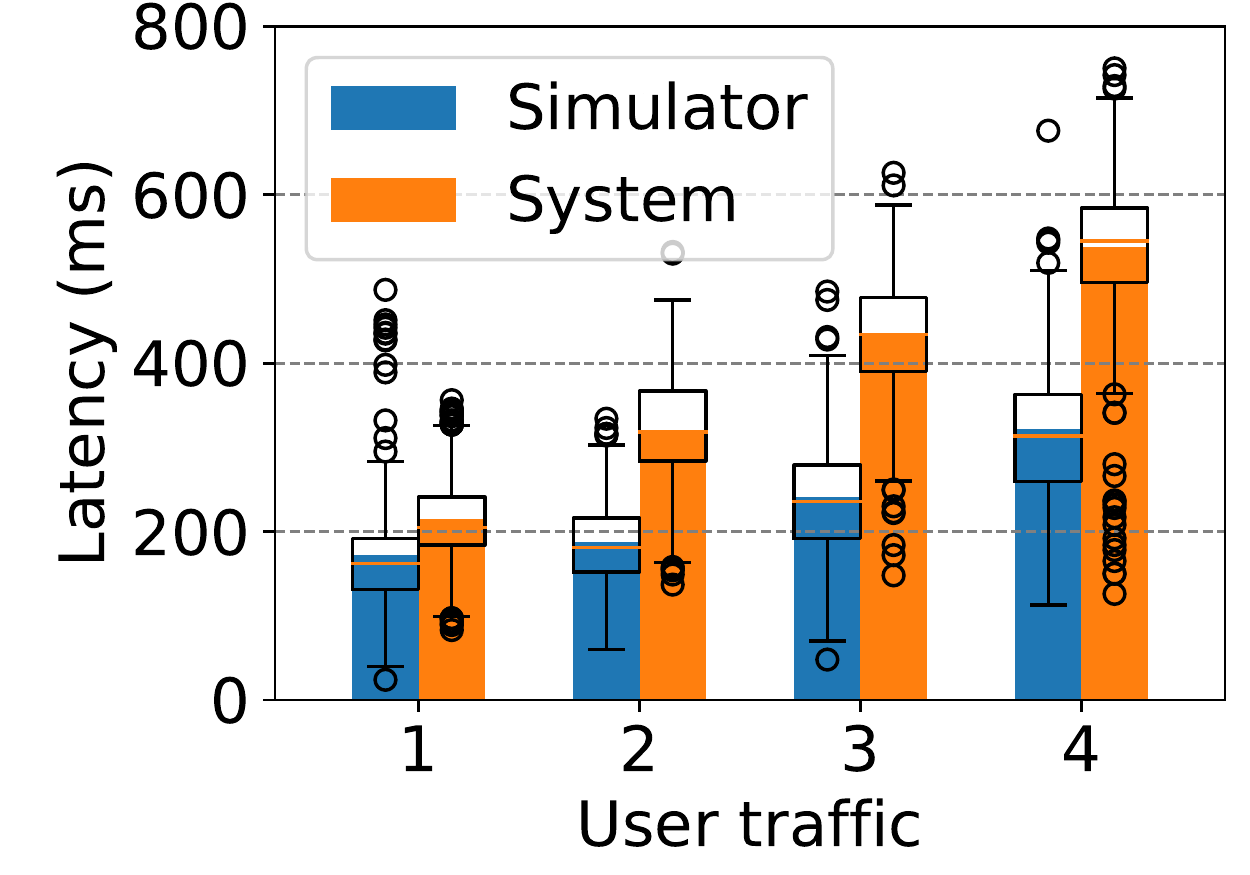}
    \captionof{figure}{\small End-to-end latency under different user traffic}
    \label{fig:result_motivation_sim_to_real_mar_traffic_comparison}
  \end{minipage}
  \begin{minipage}[t]{0.245\textwidth}
    \centering
    \includegraphics[width=1.75in, height=1.2in]{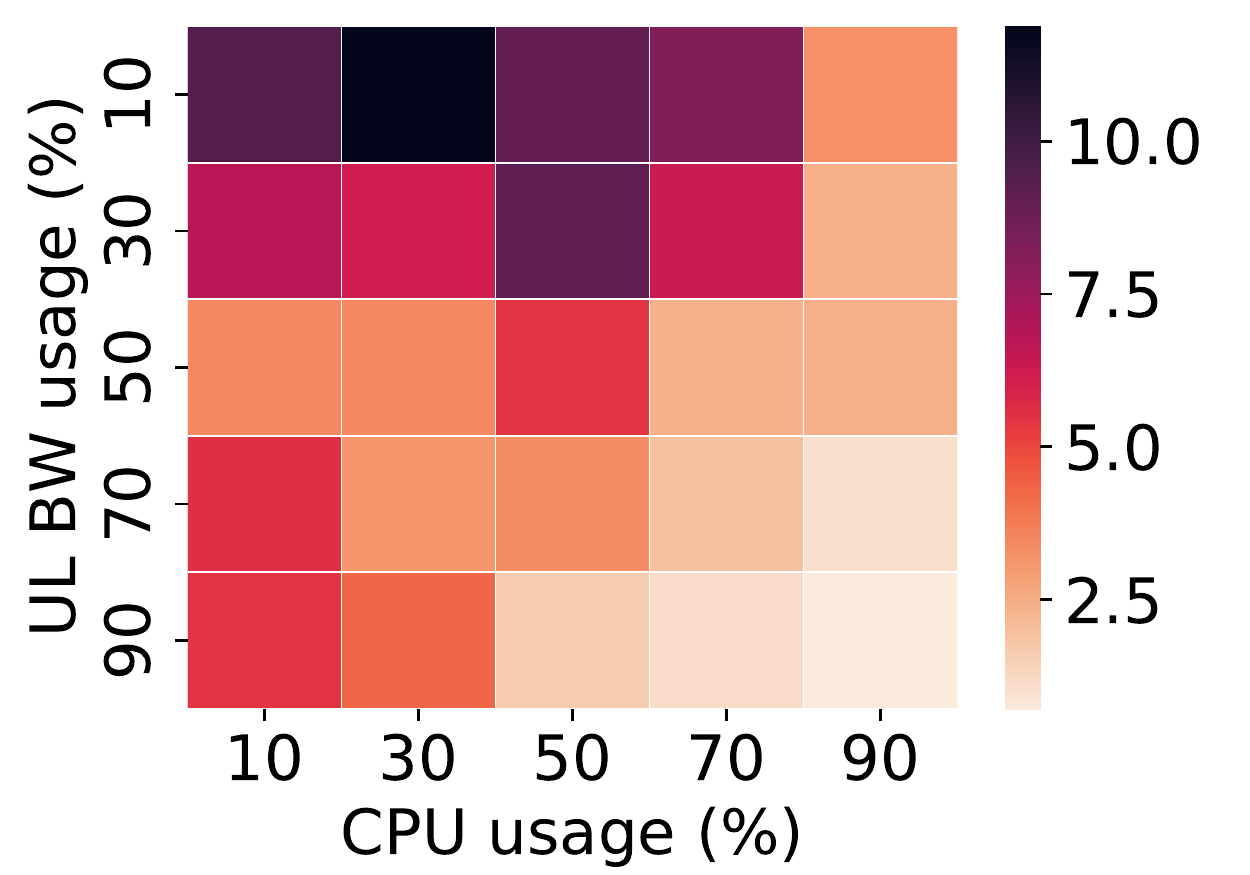}
    \caption{\small Heatmap of KL-divergence}
    \label{fig:result_motivation_sim_to_real_mar_resource_all}
  \end{minipage}
  \begin{minipage}[t]{0.245\textwidth}
    \centering
    \includegraphics[width=1.75in, height=1.2in]{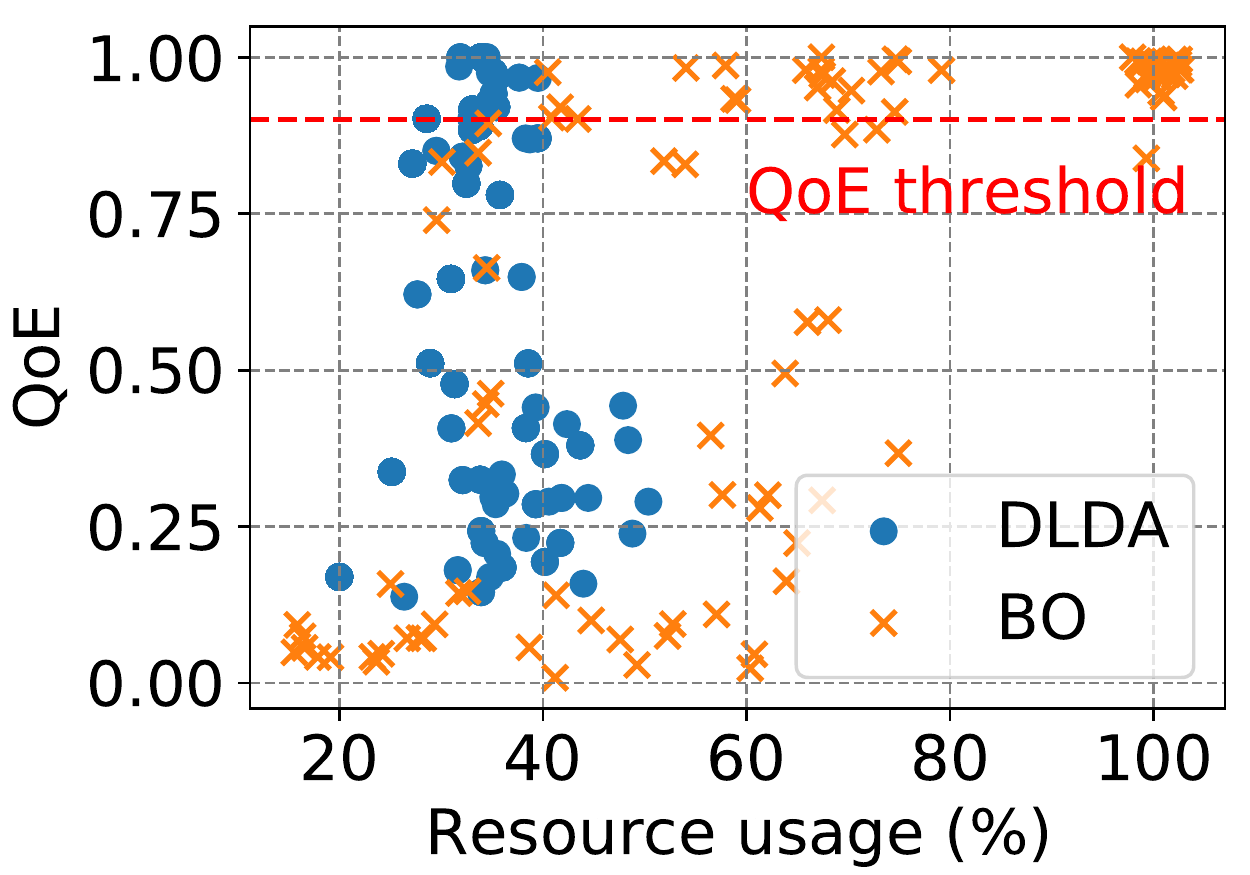}
    \captionof{figure}{\small Footprint of online learning methods}
    \label{fig:result_online_motivation_dlda}
  \end{minipage}
\end{figure*}


\vspace{-0.1in} \section{Motivation}
\label{sec:motivation}

In this section, we build a network simulator and a system prototype, evaluate the sim-to-real discrepancy in different aspects, and demonstrate the safety and sample-efficiency in online learning. 

\textbf{Setup.} 
We build a system prototype to achieve end-to-end slicing with a smartphone (OnePlus 9), an eNB (OpenAirInterface~\cite{OAI} with Ettus USRP frontend), a SDN switch (OpenDayLight~\cite{medved2014opendaylight}), and a core network (OpenAir-CN~\cite{openaircn}).
We build a network simulator by using Network Simulator 3 (NS-3)~\cite{ns3}, which includes mobile users, an eNB in radio access networks, a backhaul link in transport networks, and the EPC core networks.
A slice application is developed in both the system prototype (Android platform) and the network simulator (emulated traffic).
The simulation parameters of the simulator are matched to that of system prototype, e.g., wireless spectrum and bandwidth, user-eNB distance, application traffic and service queue.
More implementation details refer to Sec.~\ref{sec:implementation}.


\begin{table}[!t]
\small
    \begin{tabular}[b]{c|c c c}\hline
       \textbf{Performance metric}               &  \textbf{Simulator}     &  \textbf{Real Network} \\ \hline
       \textbf{ Average Ping Delay}              & 34 ms                   & 34.6 ms \\ 
       \textbf{ UL Throughput}                   & 19.87 Mbps              & 17.53 Mbps  \\ 
       \textbf{ DL Throughput}                   & 32.37 Mbps              & 31.12 Mbps \\ 
       \textbf{ UL Packet Error Rate}            & 4.16E-3                 & 9.17E-3 \\ 
       \textbf{ DL Packet Error Rate}            & 2.05E-3                 & 5.15E-3 \\ \hline
    \end{tabular}
    \captionof{table}{\small Network performance comparison (10 MHz LTE)}
\label{tb:network_performance_comparison}
\end{table}

\textbf{Sim-to-Real Discrepancy.}
We show the sim-to-real discrepancy from three perspectives, i.e., networking performances, and application performances under different user traffic and resource configurations.
First, we measure networking performances in both system and simulator in Table~\ref{tb:network_performance_comparison}.
As we can observe, the system achieves slightly lower performances in most metrics, e.g., ping delay and UL packet error rate (PER).
In particular, the UL and DL throughput of the system is 11.8\% and 3.9\% lower than that of the simulator.
These discrepancies may be attributed to a variety of factors, e.g., radio channels and completion of open-source codes. 

Second, we measure application performances under different user traffic.
Fig.~\ref{fig:result_motivation_sim_to_real_mar_traffic_1} shows the empirical cumulative probability function (CDF) of application latency when there is one user in the network, where the average latency in the system is 25.2\% higher than that in the simulator.
Besides, Fig.~\ref{fig:result_motivation_sim_to_real_mar_traffic_comparison} shows the statistics of application latency under different user traffic.
As we can see, the discrepancy between the simulator and the system becomes larger, e.g., mean and variance, when user traffic is increased.

Third, we measure application performances under different resource configurations of the slice, e.g., UL bandwidth and CPU.
Here, we use the KL-divergence~\cite{kullback1997information} to evaluate the difference between the distribution of application latency collected in the system and simulator. 
Fig.~\ref{fig:result_motivation_sim_to_real_mar_resource_all} shows that KL-divergence can be more than 10 under certain resource configurations, which can be interpreted as the two distributions are significantly different.
We notice that KL-divergence under different traffic and resource configurations are not the same, which implies uneven sim-to-real discrepancy. 

From these measurements, we observe that the sim-to-real discrepancy exists in different aspects, which is non-trivial and uneven.
The sim-to-real discrepancy could compromise the performance of offline policies in real networks, which needs to be resolved with different approaches, e.g., online learning.


\textbf{Safety in Online Learning.}
We show the performance of two state-of-the-art online learning solutions, i.e., DLDA and Bayesian Optimization (BO), in terms of safety and sample efficiency.
As DLDA~\cite{shi2021adapting} is originally designed for configuring wireless mesh networks, we modify its inputs and outputs to manage the service configuration.
Fig.~\ref{fig:result_online_motivation_dlda} shows the footprint of quality of experience (QoE\footnote{To tackle diverse performance metrics of different slices, we define a unified QoE (its value is between zero and one) to represent the slice performance for fair comparison.}) (see Eq.~\ref{prob1:const1}) and resource usage of the slice during online learning, where the QoE requirement is 0.9.
We see that both solutions can find multiple configuration actions that strike the balance between resource usage and QoE.
However, most configuration actions explored by these solutions fail to meet the QoE requirement, which needs to be avoided during online learning.
Therefore, it is imperative to design a safe and sample-efficient approach to resolve the sim-to-real discrepancy via online learning for network slicing.


\begin{figure*}
    \centering
    \includegraphics[width=6.2in, height=1.45in]{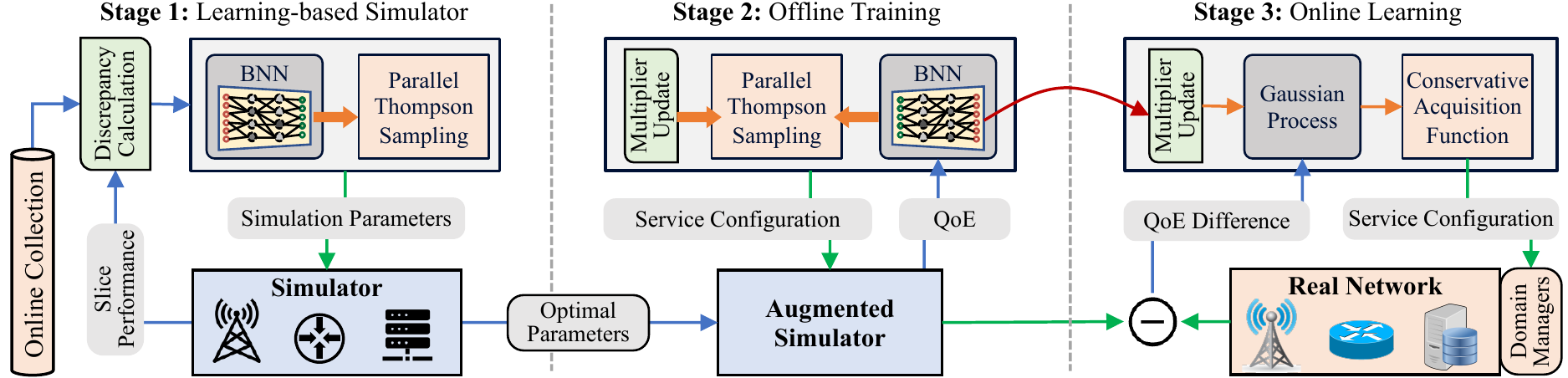}
    \caption{\small The system overview}
    \label{fig:overview}
\end{figure*}


\vspace{-0.1in} \section{System Overview}
\label{sec:system_overview}
The Atlas system includes three integrated stages, i.e., learning-based simulator, offline training, and online learning.

In the learning-based simulator stage, we use the Bayesian optimization framework to search the optimal simulation parameters for the simulator for reducing the sim-to-real discrepancy.
In particular, we design a Bayesian neural network (BNN) to approximate the complex correlation between simulation parameters and the measured sim-to-real discrepancy.
Besides, we design parallel Thompson sampling (PTS) to balance the exploration and exploitation, where the multiprocessing technique is leveraged to accelerate the searching progress.
The obtained optimal simulation parameters under the given real-world time (e.g., 1 hour), will be used in the augmented simulator in the following stages.

In the offline training stage, we aim to derive the optimal offline configuration policy to configure the cross-domain network resources to individual slices with the minimum resource usage.
To assure the QoE requirement of slices, we design an adaptive penalization method to incorporate the constraint into the objective with a dynamic multiplier.
As resource configurations are not actually implemented in real networks in this stage, intermediate violations of slice SLA are not concerned.
Hence, we train the policy based on the Bayesian optimization framework with the BNN-based approximation function, where the optimistic exploration of PTS helps to achieve better performance via offline querying the augmented simulator.
The obtained policy, e.g., the trained BNN, will serve as the start point and offline estimation in the online learning stage.

In the online learning stage, we aim to derive the optimal online configuration policy that resolves the sim-to-real discrepancy.
As every network configuration is implemented in real networks, the safety and sample efficiency become the key concerns.
Thus, we design a Gaussian process (which is sample efficient) to approximate the sim-to-real discrepancy only (which is simpler), rather than the whole correlation between resource configurations and slice QoEs.
To assure the slice SLA during intermediate explorations, we design a conservative acquisition function that achieves the Bayesian regret bound.
In addition, we design to update the multiplier in the augmented simulator, instead of based on limited online interactions, which accelerates the convergence of online learning.

\vspace{-0.1in} \section{Learning-Based Simulator Stage}
\vspace{-0.02in}
\label{sec:learning_based_simulator}
In this section, we show the design of the learning-based simulator.


\vspace{-0.1in} \subsection{The Problem}
\textbf{Network simulators.} Simulators are developed to mimic the setup of real networks by exploiting domain knowledge~\cite{omnet, ns3}, e.g., protocols, scheduling, and topology.
The simulation parameters of the simulator, e.g., link bandwidth, link delay, and pathloss model, are determined according to the corresponding specification or measurement in real networks.
However, complex network dynamics, e.g., traffic and mobility, may lead to deviated parameter values and thus compromise the accuracy of parameters.
On the other hand, abstraction mechanisms of simulators, e.g., block error rate (BLER) mapping~\cite{ikuno2010system}, do not have exact counterparts in real networks, whose parameters need to be set accurately.

\textbf{Objective.} The objective is to find the optimal simulation parameters within the given parameter space to minimize the sim-to-real discrepancy.
We consider there is an online collection $\mathcal{D}_r$ of slice performances (e.g., latency) collected from real networks\footnote{We intend to pose minimal collection efforts for network operators. For example, online collections may be collected via logging the current performance achieved by existing deployed methods.}.
Denote the offline collection $\mathcal{D}_s (\mathbf{x})$ as slice performances generated by the simulator, which is related to simulation parameters $\mathbf{x}$.
To evaluate the sim-to-real discrepancy, we resort to KL-divergence~\cite{kullback1997information} to measure the distributional differences between the two collections.
Therefore, we formulate the parameter searching problem as 
\begin{align}
\centering
&\mathbb{P}_0: &\min \limits_{ \{\mathbf{x}\}} \;\;\;\;\;\;\;\;&  {KL}[\mathcal{D}_r || \mathcal{D}_s (\mathbf{x})] \label{prob0:obj} \\ 
& &s.t.\;\;\;\;\;\;\;\;& |\mathbf{x} - \mathbf{\hat{x}}|_2 \le H, \label{prob0:const1}  
\end{align}
where ${KL}$ is the KL-divergence operator, $|\cdot|_2$ is the $l$2-norm operator, and $\mathbf{\hat{x}}$ are the original simulation parameters.
Here, we denote ${KL}[\mathcal{D}_r || \mathcal{D}_s (\mathbf{x})]$ as the sim-to-real discrepancy and $|\mathbf{x} - \mathbf{\hat{x}}|_2$ as the parameter distance.
We introduce threshold $H$ to prevent too large parameter distance, which assures the explainability of parameters with respect to original parameters. 
For instance, the sim-to-real discrepancy may be reduced to nearly zeros when all the link bandwidth are set to 10 times of the original parameter derived from technical specifications. This situation needs to be avoided, as such large parameter distance is apparently infeasible in real networks.

\textbf{Challenges.} The challenge of solving the above problem $\mathbb{P}_0$ mainly lies in the unknown (i.e., black-box) function ${KL}[\mathcal{D}_r || \mathcal{D}_s (\mathbf{x})]$, where there is no closed-form expression for modeling $\mathcal{D}_s (\mathbf{x})$.
Without the mathematical models, traditional gradient-based methods~\cite{boyd2004convex}, e.g., Newton's method, can hardly be applied.
Due to the high-dim simulation parameters, conventional searching methods (e.g., exhaustive and grid search) fails in practice, especially under the non-trivial execution time of simulators.
In other words, it is needed to design an efficient method to solve the above problem and find the optimal simulation parameters.

\vspace{-0.1in} \subsection{The Solution}
\vspace{-0.02in}
We propose an offline parameter searching algorithm based on the Bayesian optimization framework to automate the parameter searching.
Our method handles high-dim simulation parameters by using a Bayesian neural network (BNN) as the approximation function, balances exploitation and exploration via Thompson Sampling, and accelerates searching progress via parallel queries.

\textbf{Bayesian Optimization.} Bayesian optimization~\cite{frazier2018tutorial} is a state-of-the-art global optimization framework, which generally consists of probabilistic surrogate models and acquisition functions.
In each iteration, the surrogate model, e.g., Gaussian process~\cite{rasmussen2003gaussian}, is fitted with all existing observations to regress the uncertainty of the black-box function.
Then, the selected acquisition function, e.g., expected improvement (EI) and upper confidence bound (UCB)~\cite{brochu2010tutorial}, determines the utility of different candidate simulation parameters $\mathbf{x}$ for trading off exploration and exploitation.
The next simulation parameter is selected by maximizing the acquisition function, whose performance is obtained by querying the simulator.
Although Gaussian process (GP)~\cite{rasmussen2003gaussian} shows promising performance in approximating various black-box functions, its scalability~\cite{liu2020gaussian} is concerned due to its computation complexity $\mathcal{O}(n^3)$, where $n$ is the dimension of data collections.
To accurately approximate the complex correlation ${KL}[\mathcal{D}_r || \mathcal{D}_s (\mathbf{x})]$, we usually need to collect thousands or more collections from the simulator before the convergence of the Bayesian optimization.
The high-volume collections lead to ever-increasing execution time when fitting the GP model, which motivates us to explore more scalable approximation functions.

\textbf{Bayesian Neural Network.}
We design the approximation function based on Bayesian neural network (BNN), which is more scalable~\cite{snoek2015scalable} with competitive performances~\cite{jospin2022hands}, to approximate the sim-to-real discrepancy (${KL}[\mathcal{D}_r || \mathcal{D}_s (\mathbf{x})]$).
Different from standard DNNs that generate only mean-value predictions, BNN introduces stochastic components into neural network architectures, e.g., activation or weights, to quantify the uncertainty of black-box functions.
For instance, all the weights in BNNs may be represented by probability distributions, rather than having a single fixed value.

The training of BNNs aim to find the maximum a posteriori (MAP) weights expressed as $\mathbf{w}^{MAP} = \arg\underset{\mathbf{w}}{\max} \;{\log P(\mathbf{w}|\mathcal{Y})}$, where $\mathbf{w}$ are the weights of the BNN and $\mathcal{Y}$ is the collection of sim-to-real discrepancy.
Based on the Bayes' rule, the computation of posterior needs the prior $P(\mathbf{w})$ and likelihood $P(\mathcal{Y}|\mathbf{w})$, which turns out to be impractical for large neural networks.
An alternative way is the variational inference~\cite{blundell2015weight}, which approximates the complicated posterior $P(\mathbf{w}|\mathcal{Y})$ with a simpler variational approximation, e.g., Gaussian distribution.
Thus, the training of BNNs is accomplished by finding the parameter $\theta$ on a distribution on the weights $q(\mathbf{w}|\theta)$~\cite{blundell2015weight} that minimizes the KL-divergence between the true Bayesian posterior on the weights, i.e., \vspace{-0.04in}
\begin{align}
    \theta^* = & \arg\underset{\theta}{\min}\;{KL\left[ q(\mathbf{w}|\theta) || P(\mathbf{w}|\mathcal{Y})   \right]}\\ \nonumber
    = & \arg\underset{\theta}{\min}\;{KL\left[ q(\mathbf{w}|\theta)  || P(\mathbf{w})    \right] - \mathbb{E}_{q(\mathbf{w}|\theta)}\left[ \log P(\mathcal{Y}|\mathbf{w})\right]}. \vspace{-0.04in}
\end{align}
Although minimizing the above function is computationally prohibitive, \emph{Bayes-by-Backprop}~\cite{blundell2015weight} leverages the trick of re-parameterization and achieve the approximated loss as \vspace{-0.04in}
\begin{equation}
    Loss \approx \sum\nolimits_{i=1}^{N}{\log q(\mathbf{w}^{i}|\theta) - \log P(\mathbf{w}^{i}) - \log P(\mathcal{Y}|\mathbf{w}^{i})}, \vspace{-0.04in}
\end{equation}
where $\mathbf{w}^{i}$ denotes the Monte Carlo sample drawn from the variational posterior $q(\mathbf{w}^{i}|\theta)$.


\textbf{Parallel Thompson Sampling.}
Provided the BNN-based approximation function, we examine conventional acquisition functions, e.g., expected improvement (EI) and probability improvement (PI), and find that the expensive BNN prediction results in time-consuming maximization of acquisition function~\cite{wilson2018maximizing}.
The prediction of BNN is accomplished by Monte Carlo sampling (e.g., tens of duplicate inferences), where a small number of samples fail to provide accurate uncertainty estimations.

To this end, we design parallel Thompson sampling (PTS) to address this issue and balance exploration and exploitation in the Bayesian optimization.
Thompson sampling~\cite{russo2018tutorial, chapelle2011empirical} is a heuristic, effective and robust method, which samples the approximation function and selects the next querying point with the maximum utility.
Here, we extend Thompson sampling to work with BNN-based approximation functions, and achieve parallel offline queries by leveraging the multiprocessing technique.
Our basic idea is to draw an estimation of the black-box function (i.e., ${KL}[\mathcal{D}_r || \mathcal{D}_s (\mathbf{x})]$) by inferring the BNN only once, rather than evaluating the uncertainty via Monte Carlo sampling.
To determine the next simulation parameters, we randomly sample tens of thousands of simulation parameters in the given parameter space (Eq.~\ref{prob0:const1}).
After the one-time BNN inference on these samples, we select the next simulation parameters with the minimum weighted sim-to-real discrepancy (see following explanations), and query its actual performance in the simulator accordingly.
Besides, PTS achieves parallel queries by selecting multiple simulation parameters and uses multiprocessing techniques to query the simulator simultaneously.
With the parallel queries in PTS, the offline parameter searching can achieve better and more stable convergence performance (see Fig.~\ref{fig:result_simulator_parallel}).

\textbf{Weighted Sim-to-Real Discrepancy.} 
Although we limit the parameter space (i.e., Eq.~\ref{prob0:const1}), we still prefer to tradeoff the reduction of sim-to-real discrepancy and the modification of simulation parameters.
For example, we prefer to select the simulation parameters with a smaller parameter distance among these candidates who are with the same sim-to-real discrepancy.
Hence, we design to penalize the objective with a weighted parameter distance, i.e., the objective in Eq.~\ref{prob0:obj} is rewritten as ${KL}[\mathcal{D}_r || \mathcal{D}_s (\mathbf{x})]  +  \alpha |\mathbf{x} - \mathbf{\hat{x}}|_2$ (i.e., weighted discrepancy), where $\alpha$ is a non-negative weighting factor. The constraint of parameter space (i.e., Eq.~\ref{prob0:const1}) remains.

\textbf{Remark.}
In this stage, we design the learning-based simulator and propose a new parameter searching method (summarized in Appendix~\ref{sec:appendix:simulator}) that efficiently finds the optimal simulation parameters to reduce the sim-to-real discrepancy. 
The augmented simulator (with the optimal simulation parameters) serves as the offline environment for policy training in the following stages.

\vspace{-0.1in} \section{Offline Training Stage}
\vspace{-0.02in}
\label{sec:offline_training}

In this section, we present the design of network configuration policy via offline training in the augmented simulator.




\vspace{-0.1in} \subsection{The Problem}
\vspace{-0.02in}
\textbf{Network slicing.}
Consider a mobile network operator aims to support a new slice requested by a slice tenant\footnote{Atlas focuses on service configuration of individual slices, and can be extended to multiple slices scenarios because of the performance and functional isolation.}, where the slice requires multiple dedicated virtual network resources in different technical domains, e.g., RAN, TN, and edge computing.
The slice tenant makes a service-level agreement (SLA) with the network operator, where the SLA defines several key requirements of its service performance, e.g., latency, reliability and availability.
We consider the network can be discretely configured in a time-slotted manner~\cite{marquez2018should}, e.g., every hour.
As large configuration intervals weaken the temporal dependencies among consecutive configurations, the slice performances are mainly dependent on the current configurations~\cite{shi2021adapting}.
The network operator can obtain network states (e.g., user traffic) and service performance of the slice at the beginning and end of each configuration interval, respectively. 

\textbf{Objective.} We aim to derive the optimal offline policy that minimizes the network resource usage under the performance constraint of the slice by dynamically configuring the slice's resources.
Denote $\mathbf{a}_t$ as the network configuration of the slice at the time $t$ (e.g., UL and DL bandwidth), $\mathbf{s}_t$ as the network state (e.g., traffic), and $y(\mathbf{a}_t|\mathbf{s}_t)$ as the service performance, where $y(\cdot)$ is the unknown performance function.
Therefore, we formulate the network configuration problem at the time $t$ as follow\vspace{-0.05in}
\begin{align}
\centering
\label{prob_offline_training}
&\mathbb{P}_1: &\min \limits_{ \phi} \;\;\;\;\;\;\;\;&  F(\phi) \\[-3pt]
& &s.t.\;\;\;\;\;\;\;\;& Pr(y_\phi(\mathbf{a}_t|\mathbf{s}_t) \ge Y) \ge E, \label{prob1:const1}  \\[-1pt]
& &\;\;\;\;\;\;\;\;\;\;&  0 \le \mathbf{a}_t \le A, \label{prob1:const2}
\end{align}
where $\phi$ is the policy and $A$ is the maximum allowable configurations, e.g., total bandwidth.
The resource usage function $F(\phi) = |\mathbf{a}_t / A|_1$, where $|\cdot|_1$ is $l$1-norm, is developed to combine different kinds of resources, without loss of generality.
The constraints~\cite{salvat2018overbooking} in Eq.~\ref{prob1:const1} assure that the service performance of slices are better than the predefined threshold $Y$ with a higher probability $E \in [0, 1]$.

\textbf{Quality of Experience.}
We denote $Q_s(\phi) = Pr(y_\phi(\mathbf{a}_t|\mathbf{s}_t) \ge Y)$ as the QoE of the slice obtained in the augmented simulator.
Hence, the slice SLA is assured only if its requirement $E$ is satisfied in Eq.~\ref{prob1:const1}.

\textbf{Challenges.} The challenges of resolving the above problem $\mathbb{P}_1$ lie in two aspects.
First, the QoE function $Q_s(\phi)$ is unknown, where offline service performance of slices can only be obtained by executing the simulator.
Second, the QoE requirement needs to be maintained under different states $\mathbf{s}_t$, where the weighting method with fixed weights fails to achieve this constraint.

\vspace{-0.1in} \subsection{The Solution}
\label{sec:offline_training_solution}
We propose an offline network configuration algorithm to automatically learn to configure while assuring the slice SLA.
First, we design to adaptively incorporate the constraint into the objective based on the Lagrangian primal-dual method~\cite{boyd2004convex}.
Second, we leverage the Bayesian optimization framework to solve the relaxed problem, where BNN is used to approximate the unknown QoE function and parallel Thompson sampling is also exploited.

\textbf{Adaptive Penalization Method.}
The idea is to convert the constrained problem $\mathbb{P}_1$ into an unconstrained problem, by adaptively penalizing the objective with the weighted constraint.
To use the Lagrangian primal-dual method, we first build Lagrangian~\cite{boyd2004convex} as
\begin{equation}
    \mathcal{L} (\mathbf{a}_t, \lambda) = F(\phi) - \lambda (Q_s(\phi) - E), \label{eq:lagrangian}
\end{equation}
where $\lambda$ is the multiplier.
Then, the problem is resolved by alternatively solving the primal problem written as $\mathbf{a}_t^* = \arg\min\limits_{\mathbf{a}_t}  \mathcal{L}(\mathbf{a}_t, \lambda)$, and the dual problem $\mathbf{\lambda}^* = \arg\min \limits_{ \lambda \ge 0} \mathcal{L}(\mathbf{a}_t, \lambda)$.
The dual problem is solved by updating the multiplier with sub-gradient descent~\cite{boyd2004convex} as
\begin{equation}
    \lambda = \left[\lambda - \varepsilon \left( Q_s(\phi) - E \right)\right]^+,
    \label{eq:update_multiplier}
\end{equation}
where $[x]^+ = max(x,0)$ and $\varepsilon$ is a positive step size.
In this method, the multiplier $\lambda$ is increased if the slice SLA is violated, which guides the optimization in the next round.

\textbf{Learn to Configure.}
To tackle the unknown QoE function in the primal problem, we leverage the Bayesian optimization framework to obtain the optimal offline configuration policy.
In particular, a BNN is created to approximate the unknown QoE function $Q_s(\phi)$, where its inputs include the network state $\mathbf{s}_t$, threshold $Y$ and network configuration $\mathbf{a}_t$.
The exploitation and exploration are balanced with Thompson sampling, where parallel querying applies to accelerate the convergence.

\textbf{Remark.}
In this stage, we design the offline network configuration algorithm  (summarized in Appendix~\ref{sec:appendix:offline}) that derives the optimal offline policy to automatically learn to configure while assuring the slice SLA.
The derived offline policy serves multiple purposes in the online learning stage, e.g., start point and offline acceleration.

\vspace{-0.1in} \section{Online Learning Stage}
\vspace{-0.02in}
\label{sec:online_learning}

In this section, we present the design of network configuration policy via online learning to resolve the sim-to-real discrepancy.

\vspace{-0.1in} \subsection{The Problem}
\vspace{-0.02in}
The objective is to obtain the optimal policy to minimize resource usage while meeting the slice SLA, which is the same as that in the offline training stage.
The key difference is that, instead of interacting with the simulator, network configurations are queried directly to real networks in this stage.
As a result, the policy safety (i.e., maintaining the slice SLA for every network configuration) and sample efficiency (i.e., the number of needed online transitions for convergence) become two critical considerations.
For example, any unsatisfied QoEs in this stage are actually applied to actual slice users, and thus result in SLA violations in real networks.
Besides, the algorithm is highly desired to be sample efficient in real networks, e.g., 1K transitions need more than 40 days to collect when the configuration interval is 1 hour.
To evaluate the policy safety, we define the regret function with respect to the resource usage $g_n^{(u)}$ and slice QoE $g_n^{(p)}$ at the $n$th iteration as \vspace{-0.04in}
\begin{align}
\centering
g_n^{(u)} &= \sum\nolimits_{j=0}^{n} \left[F(\phi^j) - F(\phi^*) \right], \\[-2pt] 
g_n^{(p)} &= \sum\nolimits_{j=0}^{n} \left[max(Q(\phi^*) - Q(\phi^j), 0) \right],  \vspace{-0.04in}
\end{align}
where $\phi^*$ is the optimal policy, $\phi^j$ is the policy at the $j$th iteration, and $Q(\phi)$ is the slice QoE in real networks.

\vspace{-0.1in} \subsection{The Solution}
We propose an online network configuration algorithm to continue learn to configure during the online learning stage, which achieves safe exploration and sample efficient approximation function.
Specifically, we design the algorithm with an efficient Gaussian process model for regressing the sim-to-real performance difference, a conservative acquisition function with a Bayesian regret bound, and a offline acceleration method by using the offline simulator.

\textbf{Learn Sim-to-Real Discrepancy.}
Gaussian process (GP) is a widely-adopted and generic model for function approximation, which constructs probabilistic models with a variety of kernel functions to regress given data collections.
It is much more powerful than traditional parametric models, e.g., linear regression, and more sample efficient than deep neural networks (DNNs)~\cite{rasmussen2003gaussian}. 

Hence, we propose to create a GP model\footnote{The total number of online transitions is limited to hundreds, which alleviates the concern of GP scalability.} to approximate the sim-to-real performance difference only, i.e., the gap of slice QoE between the simulator and real networks.
This is based on two observations.
First, a GP model is insufficient to approximate the complicated QoE function of the slice as a whole (see Fig.~\ref{fig:result_online_convergence_usage} and Fig.~\ref{fig:result_online_convergence_qoe}).
Second, the sim-to-real performance difference is easier to be learned, provided that the BNN-based approximation function in the offline training stage has been trained extensively in the augmented simulator.
By using a GP model to approximate the sim-to-real performance difference, the slice QoE function in real networks is written as 
\begin{equation}
\label{eq:perf_real_equal_sim_plus_gap}
    Q(\phi) =  Q_s(\phi) + G(\psi),
\end{equation}
where $Q_s(\phi)$ is the slice QoE obtained in the simulator, and $G(\psi)$ is the QoE difference learned by the GP model (denoted by $\psi$).

\textbf{Conservative Exploration.}
Existing acquisition functions (e.g., EI and PI) and Thompson sampling practically cause over-exploration and lead to intermediate SLA violations (see Fig.~\ref{fig:result_online_acq_function}).
Gaussian process upper confidence bound (GP-UCB)~\cite{srinivas2009gaussian}, as an acquisition function, achieves a sub-linear regret bound with strong theoretical convergence guarantees by using a hyperparameter $\beta_t$ to balance exploration and exploitation. 
However, the hyperparameter is selected to be large to meet the regret bound requirement, which usually leads to excessive violations of slice SLA during the online learning stage. 

To this end, we propose a clipped randomized GP-UCB (cRGP-UCB)~\cite{berk2020randomised} as the acquisition function to assure conservative exploration while guaranteeing the Bayesian regret bound.
Specifically, cRGP-UCB evaluates the utility of configurations via $\mu_t(\mathbf{a}_t) + \sqrt{ \beta_t} \cdot \sigma_t(\mathbf{a}_t)$, where $\mu_t(\cdot)$ and $\sigma_t(\cdot)$ is the mean and standard deviation function, respectively.
Note that $\mu_t(\cdot)$ and $\sigma_t(\cdot)$ are estimated by using both offline BNN and online GP model (Eq.~\ref{eq:perf_real_equal_sim_plus_gap}). 
The hyperparameter $\beta_t$ is sampled from a distribution, instead of calculating to be fixed in GP-UCB, at each iteration. 
The hyperparameter is obtained by $\beta_t = \Gamma \left(\kappa_t, \rho \right)$, where $\Gamma$ is the Gamma distribution, and \vspace{-0.04in}
\begin{equation}
\label{eq:conservative_acq}
    \kappa_t = {\log\left({(n^2 + 1)}/{\sqrt{2\pi}} \right)}/{\log\left(1 + {\rho}/{2} \right)}, \vspace{-0.04in}
\end{equation}
and $\rho$ is a scaling parameter.
The distributional hyperparameter $\beta_t$ allows greater freedom to select smaller $\beta_t$, as compared to GP-UCB, which contributes to maintaining the Bayesian regret bounds\footnote{The description of Bayesian regret bounds are omitted for saving space, more details refer to~\cite{srinivas2009gaussian, berk2020randomised}.}.
As the sampled hyperparameter still can reach up to hundreds, we clip the hyperparameter for conservative exploration in practice. 
In other words, the actual $\beta_t$ is expressed as $\beta_t = clip(\beta_t, 0, B)$, where $B$ is the upper bound of hyperparameter. 
Here, both $\rho$ and $B$ can be adjusted by individual slice tenants to tradeoff potential performance improvements and possible risks of SLA violations.

\textbf{Offline Acceleration.}
To maintain the slice SLA, we use the adaptive penalization method (see Sec.\ref{sec:offline_training_solution}) to dynamically penalize the objective in Eq.~\ref{prob_offline_training} for both the offline training and online learning stage.
In online learning, the Lagrangian in Eq.~\ref{eq:lagrangian} is rewritten as  
\begin{equation}
    \mathcal{L} (\mathbf{a}_t, \lambda) = F(\phi) - \lambda (Q_s(\phi) + G(\psi) - E). \label{eq:lagrangian_online}
\end{equation}
In the offline training stage, its low convergence rate~\cite{boyd2004convex} is hidden by using parallel queries to collect multiple slice QoEs under different network configurations in simulators.  
In the online learning stage, however, the single query in real networks causes insufficient updates on the multiplier $\lambda$.
As a result, inappropriate multipliers usually lead to changing resource usage and frequent violations of slice SLA, e.g., too small multipliers may fail to penalize the objective for assuring the slice SLA.

To this end, we propose to exploit the augmented simulator to update the multiplier for accelerating the convergence of online algorithm in the online learning stage.
Our basic idea is to update the multiplier multiple times by estimating the slice QoE with Eq.~\ref{eq:perf_real_equal_sim_plus_gap}, where the $Q_s(\phi)$ is obtained in the augmented simulator and $G(\psi)$ is predicted by the current online GP model $\psi$.
Hence, the update of multiplier in Eq.~\ref{eq:update_multiplier} is rewritten as 
\begin{equation}
    \lambda = \left[\lambda - \varepsilon \left( Q_s(\phi) + G(\psi) - E \right)\right]^+.
    \label{eq:update_multiplier_revised}
\end{equation}
In other words, the slice QoE obtained from real networks serves more to regress the GP model $\psi$.
As more online transitions are collected, a more accurate GP model can be achieved, which helps to estimate actual sim-to-real performance differences.

\textbf{Remark.}
In this stage, we design the online algorithm (summarized in Appendix~\ref{sec:appendix:online}) that derives the optimal online policy to resolve the sim-to-real discrepancy via online learning with real networks.
The policy is composed of two models, i.e., the offline BNN $\phi$ learns offline estimation of slice QoE $Q_s(\phi)$ and online GP $\psi$ learns only sim-to-real discrepancy $G(\psi)$.

\begin{figure}
    \centering
    \includegraphics[width=3.3in, height=1.8in]{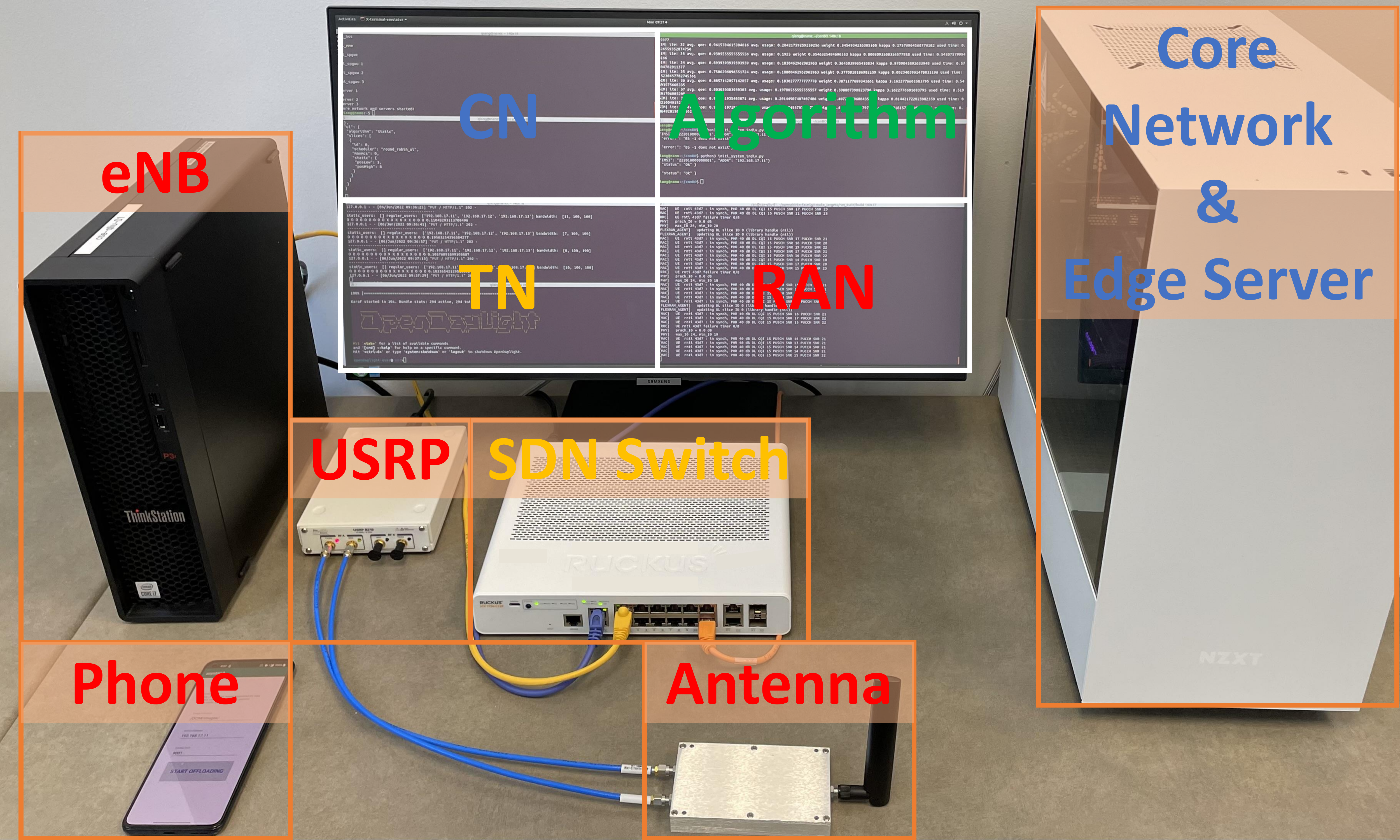}
    \caption{\small Overview of system prototype}
    \label{fig:testbed}
\end{figure}

\section{System Implementation}
\vspace{-0.02in}
\label{sec:implementation}

In this section, we present implementation of end-to-end slicing prototype (Fig.~\ref{fig:testbed}), network simulator and algorithms.

\vspace{-0.1in} \subsection{The System}
\vspace{-0.02in}
\textbf{RAN.} We implement the radio access network based on OpenAirInterface (OAI)~\cite{OAI} with 4G LTE.
The eNB operates at band 7 with 10MHz radio bandwidth, i.e., 50 physical resource blocks (PRBs).
The RAN is hosted in an Intel i7 desktop with a low-latency kernel of Ubuntu 18.04, which connects an Ettus USRP B210 as the RF front-end.
The distance between the eNB antenna and stationary smartphone is 1 meter.
To enable the network slicing capability, we develop the radio domain manager with FlexRAN support~\cite{foukas2016flexran}.

\textbf{UEs.}
We use a OnePlus 9 5G smartphone (Qualcomm Snapdragon 888 and Android 11) as the user to connect with the eNB.
We develop an Android application\footnote{Atlas makes no prior assumptions on the performance metric of slices and is compatible with other applications.} for the smartphone, which continuously sends frames (540p) to the edge server in CN.
The server processes the frame with a feature extraction algorithm (ORB~\cite{rublee2011orb}) and then feeds the results back.
We limit the number of on-the-fly frames (i.e., the frames with no results back yet) for the purpose of congestion control.
The performance metric of this application is end-to-end latency.
To emulate varying user traffic, we control the number of on-the-fly frames, e.g., we may set it as four to emulate the traffic from four users.


\begin{table}[!t]
\small
    \begin{tabular}[b]{c|c|c}\hline
       \textbf{Configuration}                       &  \textbf{Meaning} & \textbf{Range} \\ \hline
       \textbf{ \emph{bandwidth\_ul}}        &  maximum uplink PRBs  & [0, 50]\\ 
       \textbf{ \emph{bandwidth\_dl}}           & maximum downlink PRBs & [0, 50] \\ 
       \textbf{ \emph{mcs\_offset\_ul}}      & uplink MCS offset~\cite{liu2021onslicing} & [0, 10]\\     
       \textbf{ \emph{mcs\_offset\_dl}}       & downlink MCS offset~\cite{liu2021onslicing} & [0, 10]\\     
       \textbf{ \emph{backhaul\_bw}}       & transport bandwidth (Mbps) & [0, 100]\\      
       \textbf{ \emph{cpu\_ratio}}       & CPU ratio of docker  & [0, 1.0]\\  \hline    
    \end{tabular}
    \captionof{table}{\small Network configuration space}
\label{tb:configuration_space}
\end{table}

\begin{table}[!t]
\small
    \begin{tabular}[b]{c|c}\hline
       \textbf{Parameters}                      &  \textbf{Meaning} \\ \hline
       \textbf{ \emph{baseline\_loss}}           &  base loss in pathloss model (dBm)  \\ 
       \textbf{ \emph{enb\_noise\_figure}}      & noise by non-ideal transceivers (dBm)  \\ 
       \textbf{ \emph{ue\_noise\_figure}}        & noise by non-ideal transceivers (dBm) \\     
       \textbf{ \emph{backhaul\_bw}}             & additional transport bandwidth (Mbps)\\     
       \textbf{ \emph{backhaul\_delay}}          & additional transport delay (ms)\\      
       \textbf{ \emph{compute\_time}}            & additional server compute time (ms)\\      
       \textbf{ \emph{loading\_time}}            & additional loading time in UE (ms)\\  \hline    
    \end{tabular}
    \captionof{table}{\small Simulation parameter space}
\label{tb:simulation_parameter_space}
\end{table}

\textbf{TN.}
We implement the transport network based on OpenDayLight (ODL)~\cite{medved2014opendaylight} with OpenFlow 1.3.
We use a Ruckus ICX 7150-C12P as the SDN switch to connect the eNB and CN, where each port has 1Gbps capacity.
To enable the network slicing capability, we develop the transport domain manager by dynamically modifying the link bandwidth with \emph{meters} API in OpenFlow protocol~\cite{mckeown2008openflow}.

\begin{table*}[!t]
\small
    \begin{tabular}[b]{c|c|c|c}\hline
       \textbf{Methods}                      &  \textbf{Sim-to-Real Discrepancy} &  \textbf{Parameter distance}  &  \textbf{Best simulation parameters} \\ \hline
       \textbf{ Original Simulator}                   & 1.38  & 0 &     [38.57, 5.0, 9.0, 0.0, 0.0, 0.0, 0.0]      \\ 
       \textbf{ Aug. Simulator, GP}            & 0.31   & 0.16  &  [38.57, 1.44, 7.48, 5.07, 9.23, 6.02, 6.47]   \\ 
       \textbf{ Aug. Simulator, Ours}        & 0.26  & 0.12  &     [38.76, 0.68, 8.93, 5.03, 8.93, 2.16, 3.10]   \\  \hline    
    \end{tabular}
    \captionof{table}{\small Details of offline learning-based simulator}
\label{tb:offline_augmented_simulator}
\end{table*}

\begin{figure*}[!t] 
\captionsetup{justification=centering}
  \begin{minipage}[t]{0.235\textwidth}
    \centering
    \includegraphics[width=1.75in, height=1.2in]{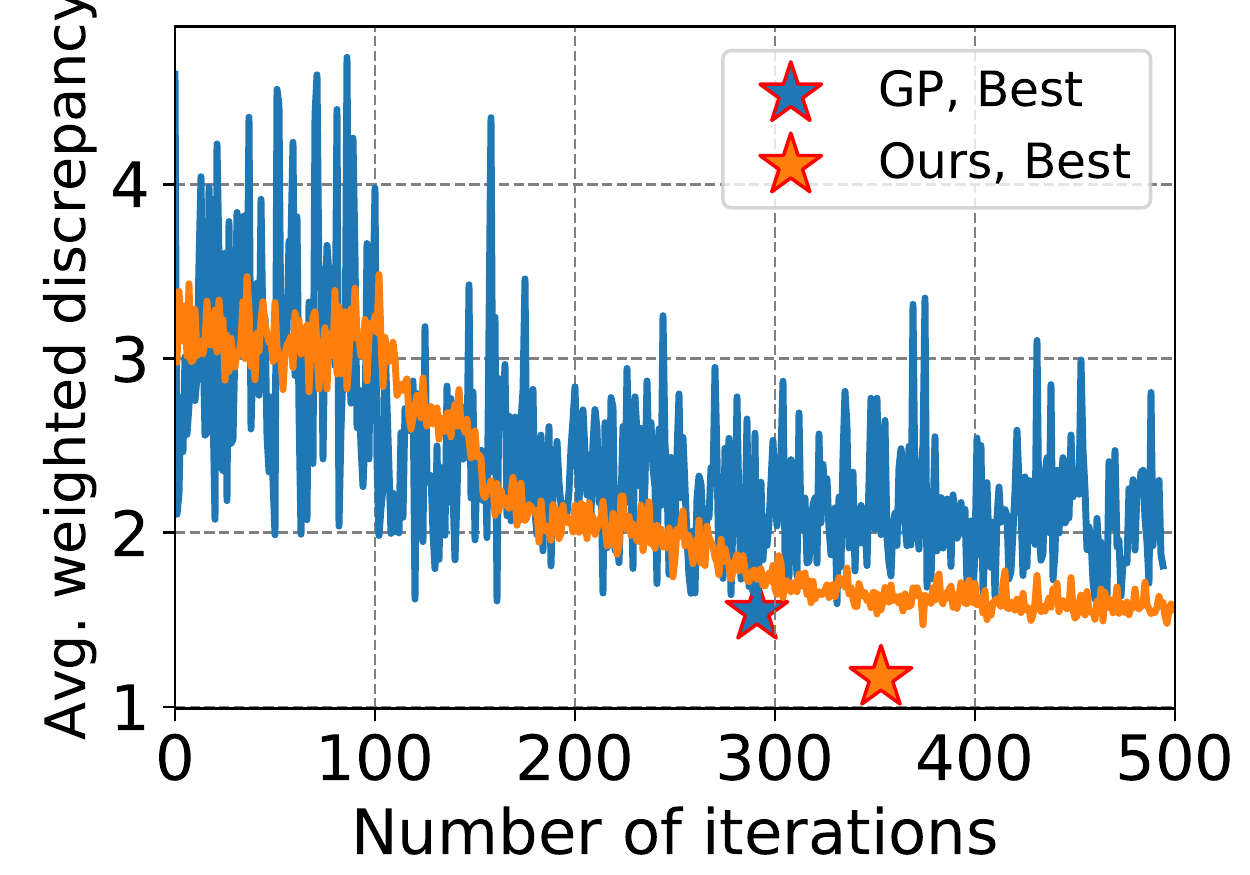}
    \caption{\small Searching progress under different methods}
    \label{fig:result_simulator_convergence}
  \end{minipage}
  \begin{minipage}[t]{0.235\textwidth}
    \centering
    \includegraphics[width=1.75in, height=1.2in]{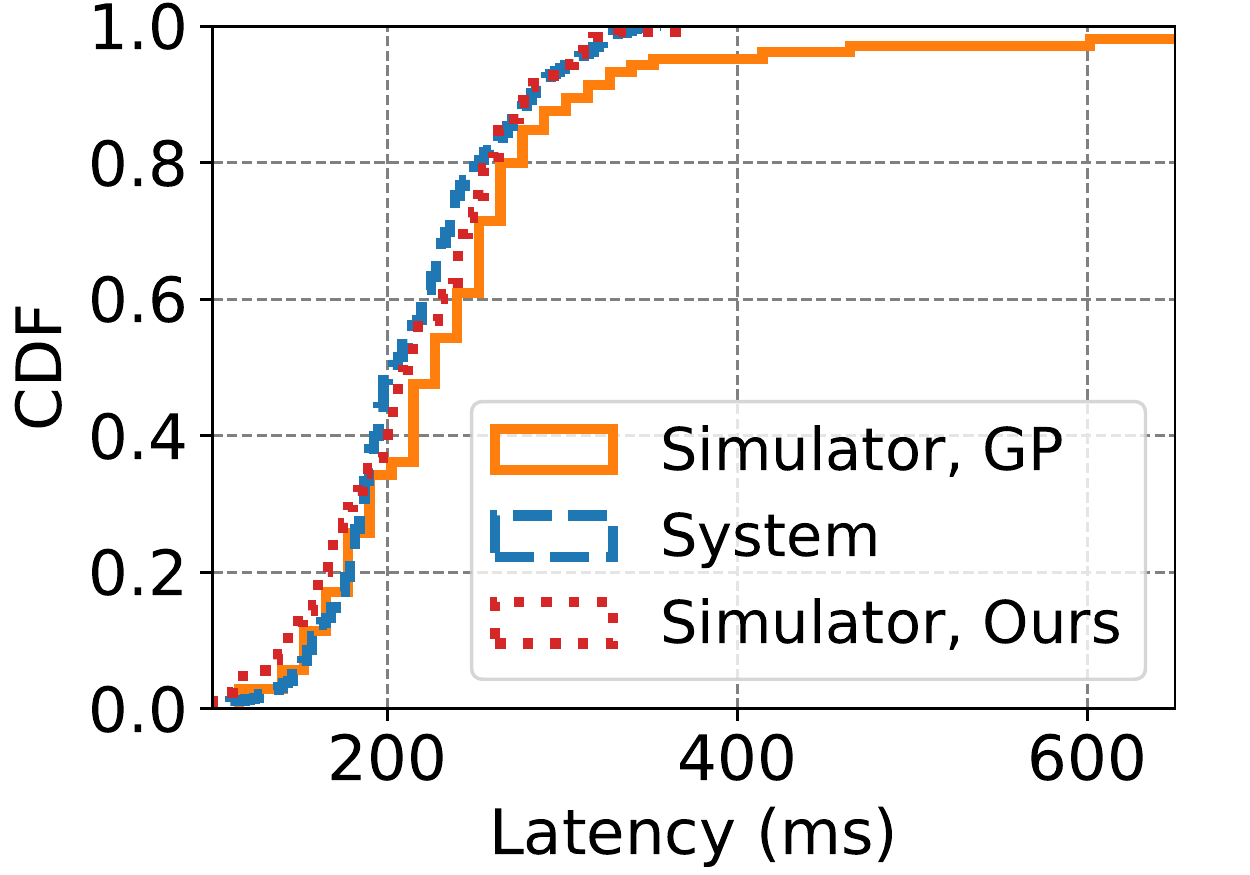}
    \captionof{figure}{\small CDF of latency under different methods}
    \label{fig:result_simulator_learned_cdf_comparison}
  \end{minipage}
  \begin{minipage}[t]{0.235\textwidth}
    \centering
    \includegraphics[width=1.75in, height=1.2in]{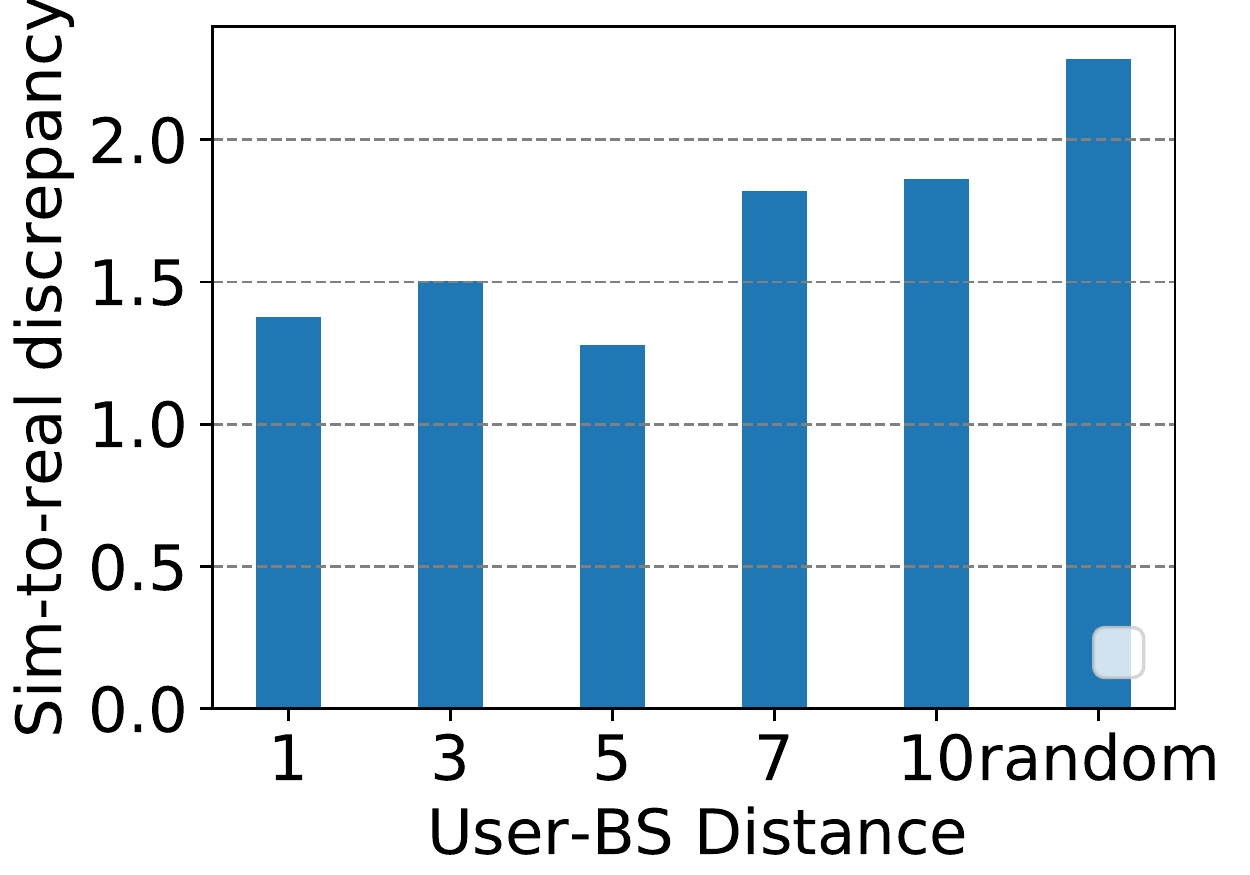}
    \caption{\small Sim-to-real discrepancy under user mobility}
    \label{fig:result_distance_user_base_station}
  \end{minipage}
  \begin{minipage}[t]{0.235\textwidth}
    \centering
    \includegraphics[width=1.75in, height=1.2in]{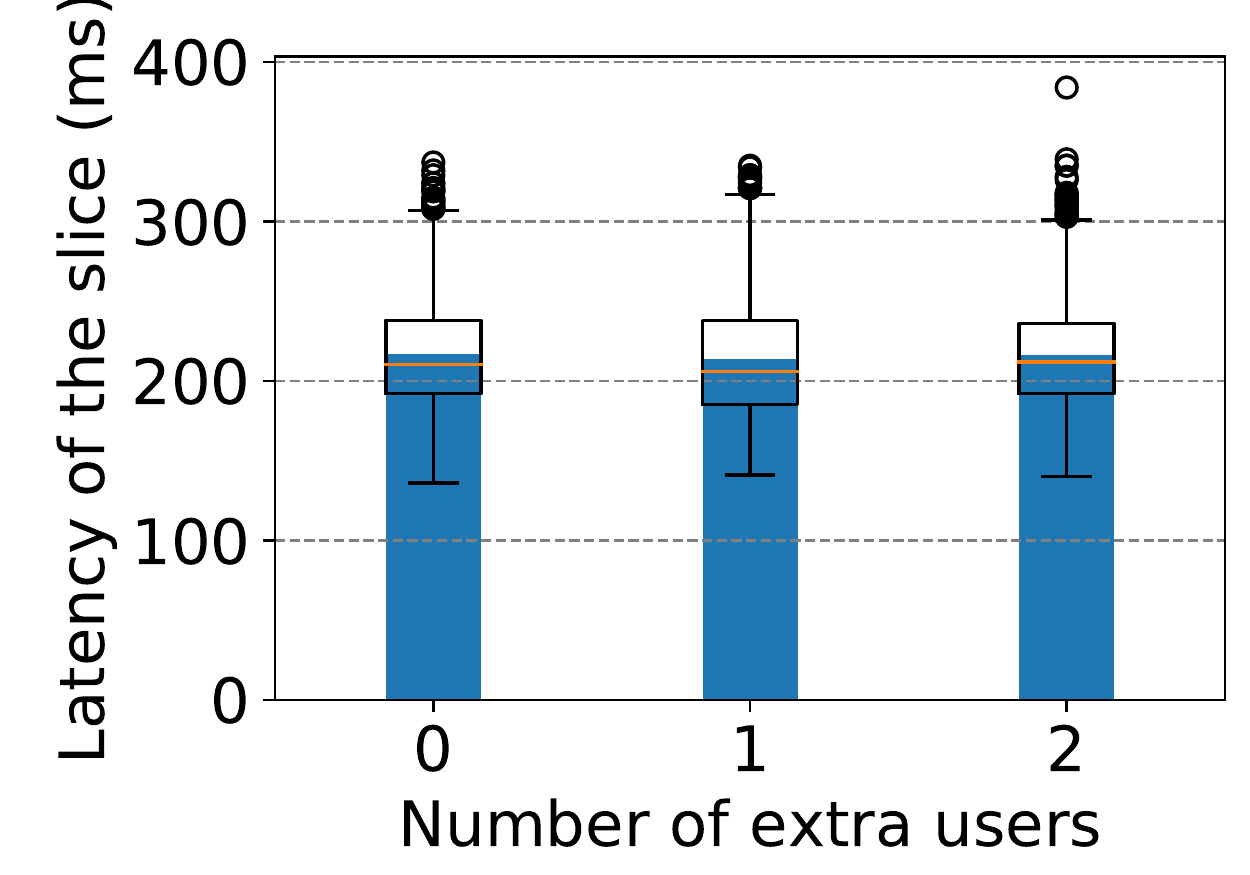}
    \captionof{figure}{\small Slice latency under extra mobile users}
    \label{fig:result_isolation}
  \end{minipage}
\end{figure*}

\textbf{CN.}
We implement the core network based on OpenAir-CN~\cite{openaircn} with the control and data plane separation architecture (CUPS).
The network functions, e.g., HSS, MME, SPGW-C, and SPGW-U, are deployed with Docker containers.
To enable the network slicing capability, we develop the core domain manager by mapping users' destination SPGW-U to that of slices.
Thus, each slice has an isolated SPGW-U container for serving its own slice users, while the other network functions are shared.

\textbf{EN.}
We implement the edge computing network based on Docker container~\cite{merkel2014docker}, which virtualizes the computing resources and provides isolation for edge servers.
For the sake of simplicity, we co-locate the edge server of slices with their SPGW-U containers.
The edge server serves the slice users, where the performances can be retrieved with a minimum 60-second interval via the developed REST API.
Besides, we develop the edge domain manager to dynamically manage the CPU ratio of edge servers, via \emph{docker update}.

\textbf{Configuration Space.}
We develop 6-dim network configuration actions in both RAN, TN and EN in the prototype, see Table~\ref{tb:configuration_space}. 

\vspace{-0.1in} \subsection{The Simulator}
\vspace{-0.02in}
The network simulator is developed based on Network Simulator 3 (NS-3)~\cite{ns3}, which is an extensively adopted platform and capable of conducting various network simulations.
We develop 7-dim simulation parameters in the NS-3 simulator, see Table~\ref{tb:simulation_parameter_space}.
The cellular network is developed based on the LENA project with 4G LTE, where we adopt the \emph{LogDistancePropagationLossModel} pathloss model and no fading model.
The transport network is simulated by a \emph{p2p} link, whose bandwidth and delay are matched to experimental measurements in the prototype.
We develop the edge computing module to allow queue-based computing simulation (matched to the prototype), where the computing delay is sampled from the experimental collections (81ms mean and 35ms std).
The Android application is also replicated in NS-3, which matches the traffic pattern and uplink transmission size (28.8kb mean and 9.9kb std).
All other settings are thoughtfully examined and configured to match that in the prototype, e.g., MAC scheduler algorithm, antenna type and gain, frequency band, and distance between eNB and smartphones.
The simulation results are obtained by reading the tracer including not only end-to-end latency of every frame, but also transmission and computing details, e.g., queuing time, computing time, and uplink and downlink transmission time.

\vspace{-0.1in} \subsection{The Algorithm}
\vspace{-0.02in}
We develop BNNs with PyTorch 1.5, where neural networks use 4-layer fully connected layers, i.e., 128x256x256x128, with \emph{ReLU} activation functions~\cite{goodfellow2016deep}.
We adopt the \emph{Adadelta} optimizer with the initial learning rate of 1.0, where the learning rate is decayed by using the \emph{StepLR} scheduler with gamma 0.999 and batch size 128.

We develop the GP model using \emph{sklearn} toolkit~\cite{scikit-learn} with the \emph{GaussianProcessRegressor} module.
We adopt the \emph{Matern} kernel with $\nu=2.5$, which is a generalization of the radial-basis function (RBF) kernel.
Besides, target values are normalized by removing the mean and scaling to unit-variance for better regression performance.

To evaluate the computation complexity of \emph{Atlas}, we use the \emph{cProfile} tool to profile our methods in a desktop with AMD Ryzen 5 3600 and 32G DDR4 RAM.
In learning-based simulator stage, we obtain the computation time of 22.27s with 214201 calls per iteration, and 0.4GB memory usage.
In the offline training stage, we obtain the computation time of 27.23s with 211665 calls per iteration, and 0.5GB memory usage.
In the online learning stage, we obtain the computation time of 16.99s with 357292 calls per iteration, and 1.1GB memory usage.




\vspace{-0.1in} \section{Performance Evaluation}
\vspace{-0.02in}
\label{sec:evaluation}

In this section, we conduct extensive network simulations and experiments to evaluate the performance of Atlas.
The step size of dual problem updates in Eq.~\ref{eq:update_multiplier_revised} is $\varepsilon=0.1$.
The scaling factor $\rho$ is 0.1~\cite{berk2020randomised} and the clipping value $B$ is 10 to prevent too large exploration.
We determine the fixed weight $\alpha$ is 7 to balance the reduction of sim-to-real discrepancy and the parameter distance.
The application related parameters, i.e., $E = 0.9$, and $Y = 300 ms$, are set according to capability of the prototype, e.g., RAN throughput.
The number of offline and online iterations is 1000 and 100, respectively.
The simulation and experimental time of each network configuration are 60 seconds, for collecting statistical performance.
We compare Atlas with the following solutions.
\begin{itemize}[leftmargin=*]
    \item \emph{Baseline.} The baseline uses the Bayesian optimization with GP model and expected improvement (EI) acquisition function to online learn in real networks directly.
    \item \emph{DLDA.} The DLDA~\cite{shi2021adapting} is a state-of-the-art online learning solution, which transfers the offline knowledge from a teacher model to the student model via interacting with real networks. As DLDA is designed to improve the prediction accuracy, we modify it to choose the configuration with minimum resource usage while meeting the QoE requirement. This is completed by randomly sampling 10K configurations\footnote{As the dimensional of the configuration space is six, we believe 10K samples are sufficient to seek the optimal action under controlled accuracy.} from the configuration space.
    \item \emph{VirtualEdge.} The VirtualEdge~\cite{liu2019virtualedge} uses a GP model to online learn the unknown slice QoE function, and relies on a predictive gradient descent method to update the current configuration under accumulative online interactions. 
\end{itemize}


\begin{figure*}[!t] 
\captionsetup{justification=centering}
  \begin{minipage}[t]{0.245\textwidth}
    \centering
    \includegraphics[width=1.75in, height=1.2in]{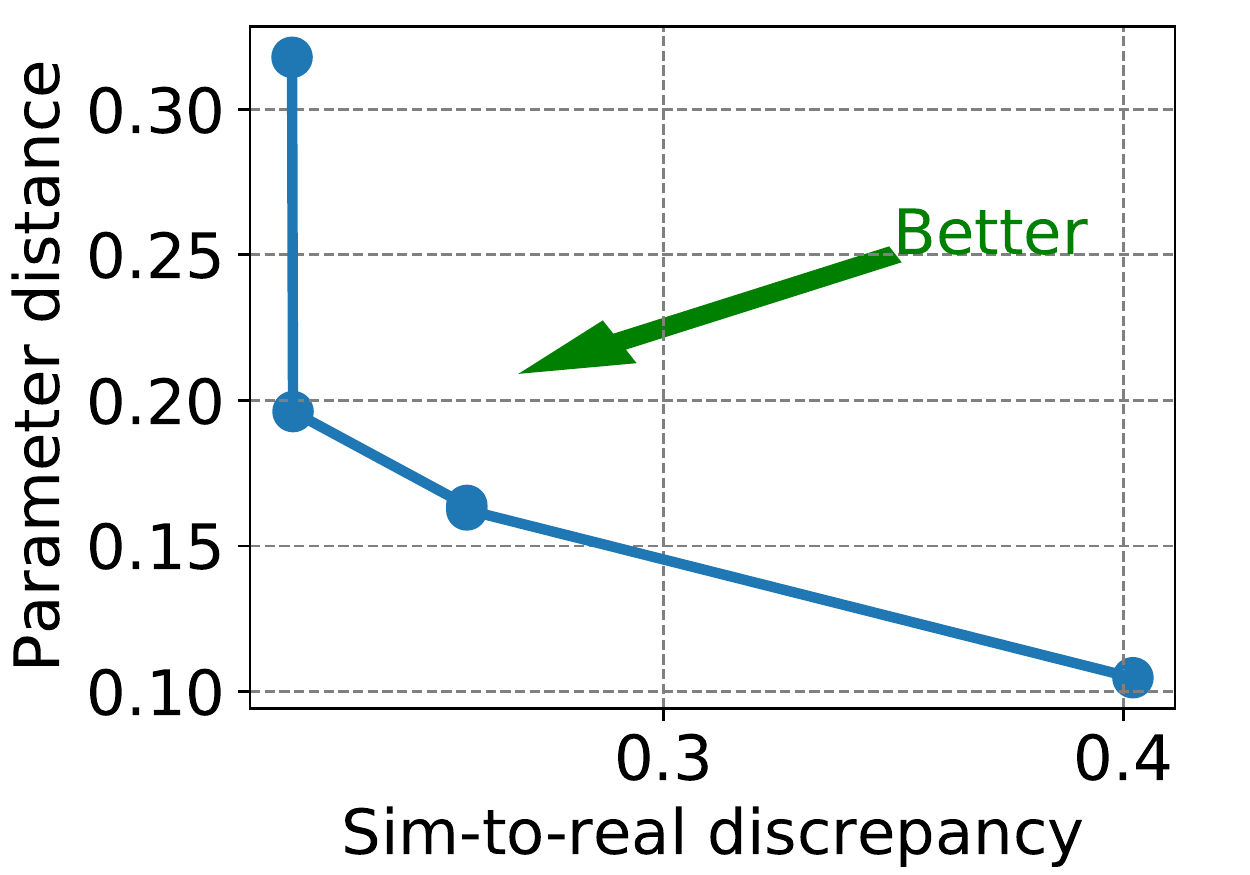}
    \captionof{figure}{\small Pareto boundary of augmented simulator}
    \label{fig:result_simulator_weight_pareto}
  \end{minipage}
  \begin{minipage}[t]{0.245\textwidth}
    \centering
    \includegraphics[width=1.75in, height=1.2in]{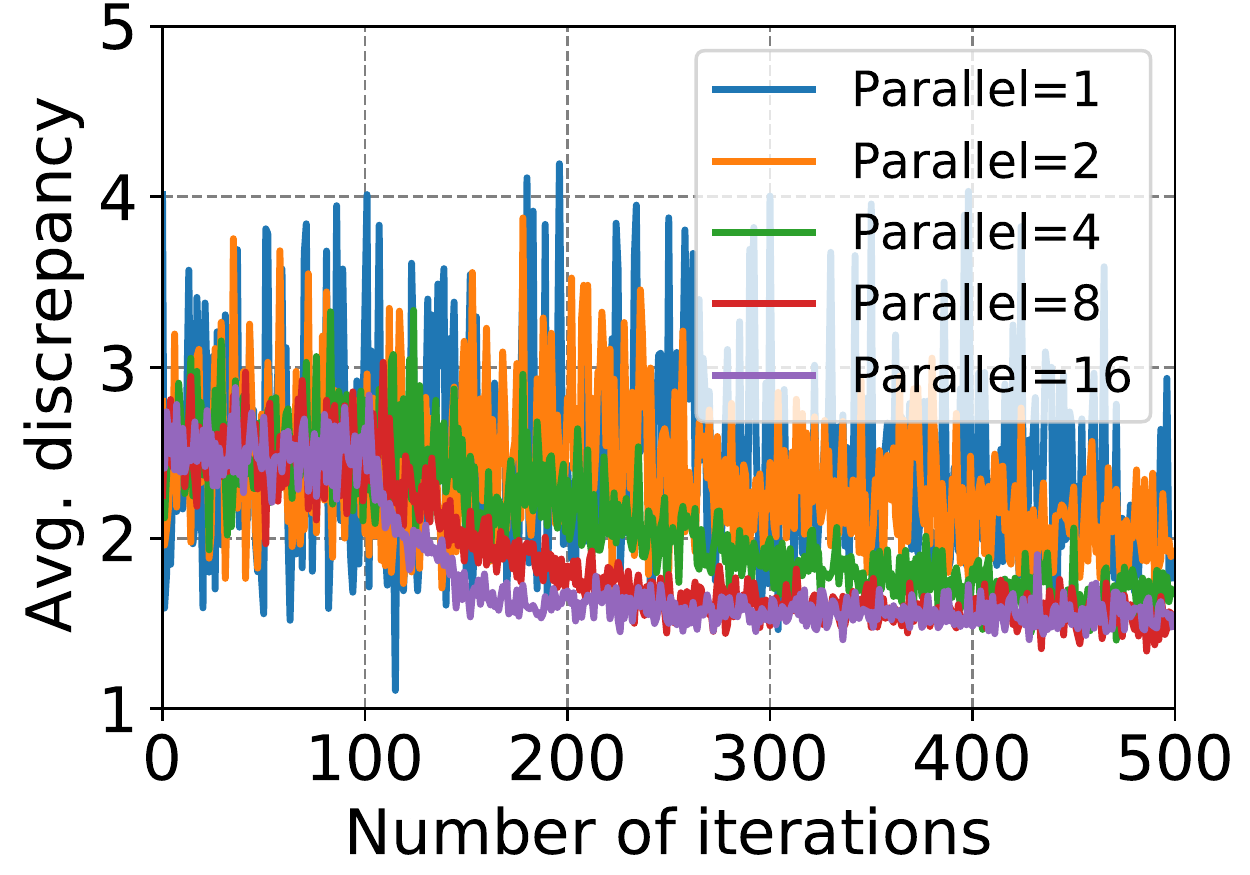}
    \captionof{figure}{\small Searching progress with parallel queries}
    \label{fig:result_simulator_parallel}
  \end{minipage}
  \begin{minipage}[t]{0.245\textwidth}
    \centering
    \includegraphics[width=1.75in, height=1.2in]{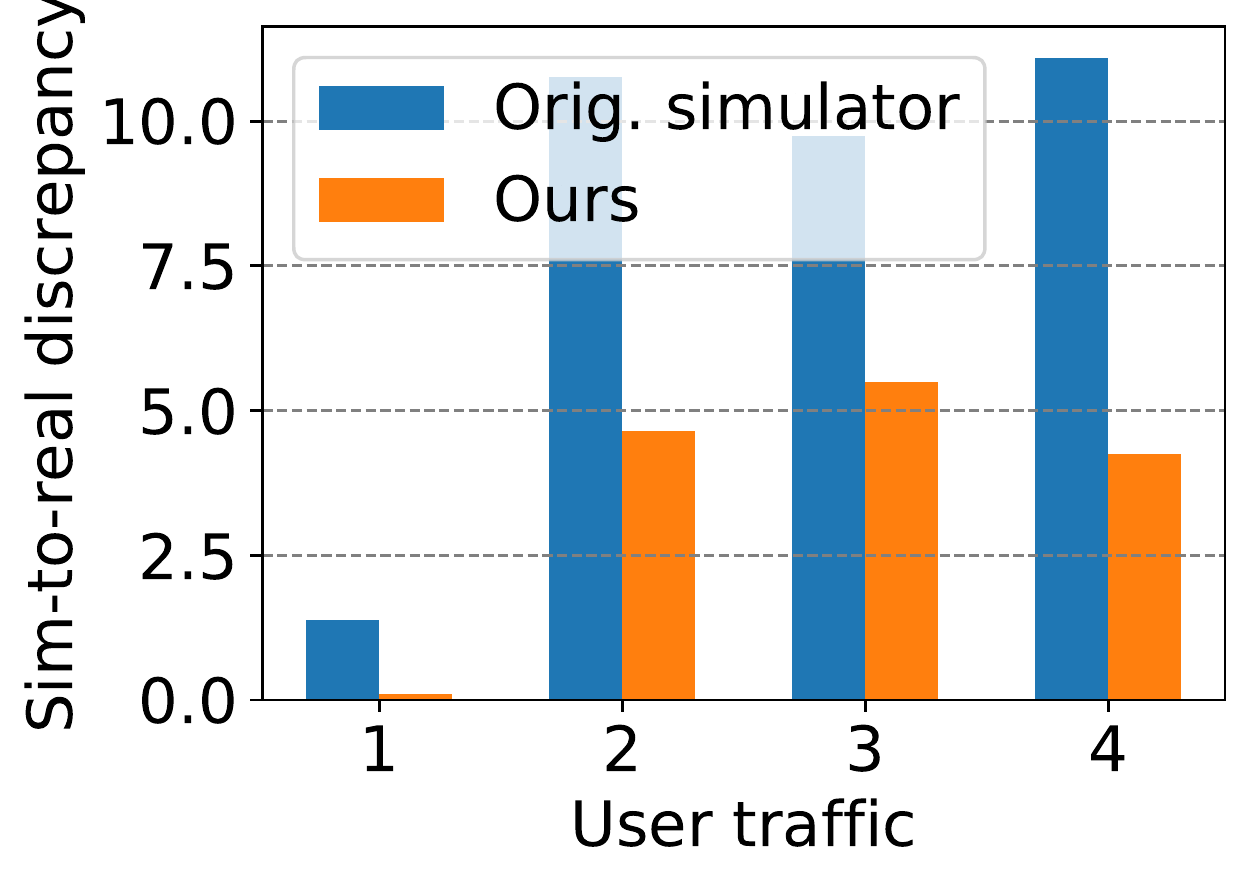}
    \captionof{figure}{\small Discrepancy reduction under user traffic }
    \label{fig:result_online_evaluation_traffic}
  \end{minipage}
  \begin{minipage}[t]{0.245\textwidth}
    \centering
    \includegraphics[width=1.75in, height=1.2in]{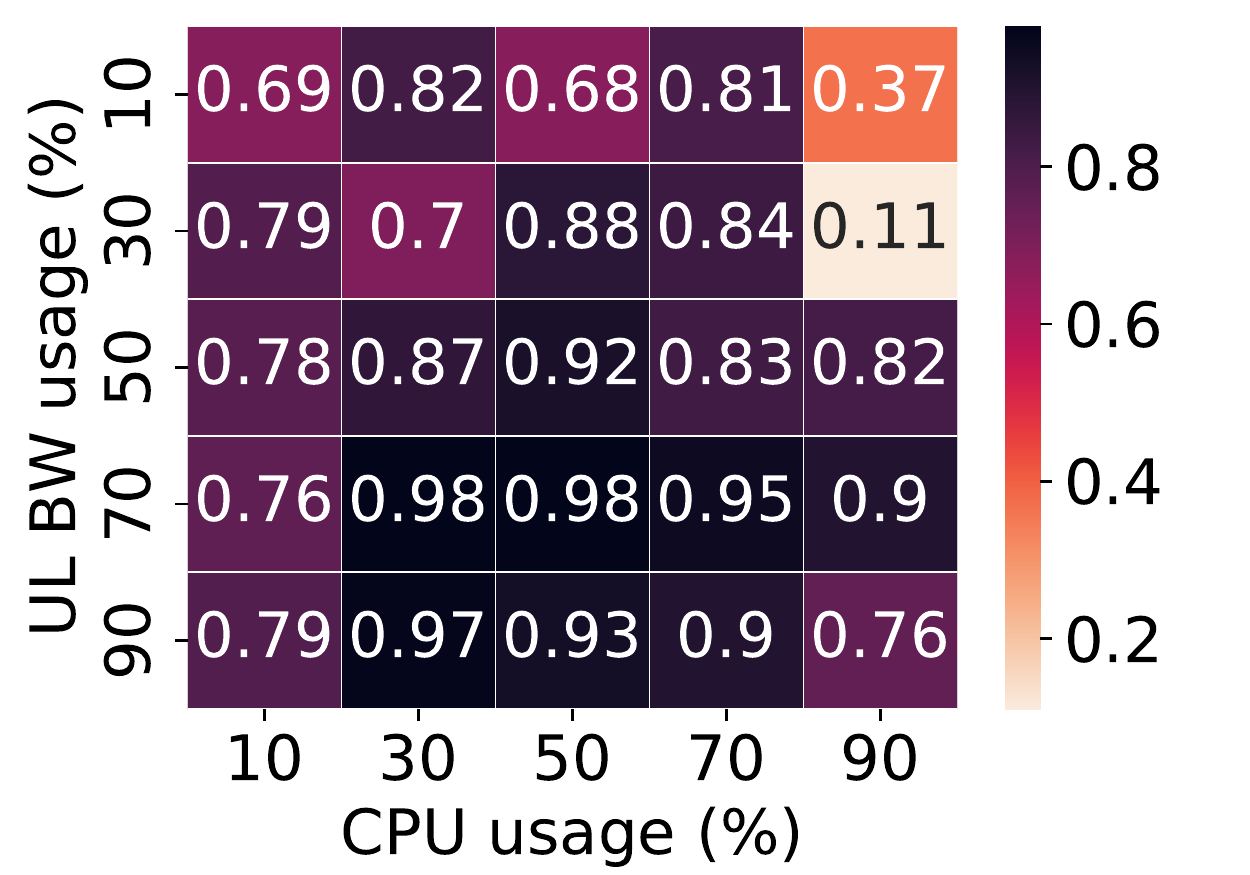}
    \captionof{figure}{\small Discrepancy reduction (1.0 means 100\%) under resources }
    \label{fig:result_motivation_sim_to_real_mar_aug_gain_resource_all}
  \end{minipage}
\end{figure*}

\vspace{-0.1in} \subsection{Learning-based Simulator}
\vspace{-0.02in}
In the learning-based simulator stage, we focus on the metric of sim-to-real discrepancy under given number of iterations.

\textbf{Searching Progress.}
Fig.~\ref{fig:result_simulator_convergence} shows the searching progress of different methods, where the first 100 iterations are purely exploration.
The average weighted discrepancy is calculated by averaging the weighted sim-to-real discrepancy in each iteration.
Our method finds better simulation parameters, which reduce 24.5\% average weighted discrepancy than that of the GP-based approach.
In the experiments, conducting a 60-second simulation in NS-3 consumes average 27.8 seconds real-world time\footnote{Leveraging the multiprocessing technique, parallel querying of NS-3 with 16 processes achieves nearly the same real-world time with single query.}.

Table~\ref{tb:offline_augmented_simulator} shows the details of best simulation parameters obtained by different methods, where the order of simulation parameters refers to Table.~\ref{tb:simulation_parameter_space}.
The original simulator uses the default simulation parameters, where \emph{ReferenceLoss} is 38.57 in the \emph{LogDistancePropagationLossModel} pathloss model in NS-3, and the UE and eNB noise figure are 9.0 and 5.0, respectively.
As a result, the original simulator has zero parameter distance, while gets 1.38 sim-to-real discrepancy.
The GP-based method obtains 0.31 sim-to-real discrepancy under 0.16 parameter distance, with heavily increased \emph{loading\_time}, \emph{backhaul\_bw} and \emph{compute\_time}.
In contrast, our method achieves 0.26 sim-to-real discrepancy (i.e., 81.2\% reduction than original simulator) with 0.12 parameter distance.
As we calculate the parameter distance with L-2 norm, the actual parameter difference is even smaller, which helps to maintain the explainability of simulation parameters.
Fig.~\ref{fig:result_simulator_learned_cdf_comparison} shows the cumulative probability of latency under the best simulation parameters obtained by different methods.
We can see that the CDF achieved by the GP-based approach has a long tail, which is worse than that of our method.
As compared to the CDF of original simulator (see Fig.~\ref{fig:result_motivation_sim_to_real_mar_traffic_1}), the sim-to-real discrepancy is substantially reduced with the learning-based simulator.

\textbf{Network Dynamics.}
Fig.~\ref{fig:result_distance_user_base_station} shows the sim-to-real discrepancy obtained under different user mobility.
In general, we see the sim-to-real discrepancy increases under a larger line-of-sight distance between the user and the base station.
This may be attributed to the disparity of the radio channel model, where the pathloss model in the simulator fails to represent the real channel dynamics in experiments, especially under random walk scenarios.
Besides, to evaluate the end-to-end performance isolation among network slices, we dynamically attach new users, generate traffic (i.e., YouTube), and detach the users.  
Fig.~\ref{fig:result_isolation} shows that the latency performance of the slice is very stable no matter how many extra users are in the network.
This is attributed to the performance isolation achieved in both RAN, TN, CN, and EN in the developed system prototype.

\textbf{Pareto Boundary.}
Fig.~\ref{fig:result_simulator_weight_pareto} shows the Pareto boundary achieved by our method via varying the weight $\alpha$.
Provided that the sim-to-real discrepancy is 0.26 when $\alpha=7$, it can be further reduced to 0.21 at the cost of more than 0.2 parameter distance.
On the other hand, given the maximum 0.1 parameter distance, the lowest sim-to-real discrepancy can reach 0.4, which still achieves a 71.0\% reduction as compared to that of the original simulator.
Besides, our method allows the customized weight to balance sim-to-real discrepancy and the parameter distance.

\textbf{Parallel Queries.}
Fig.~\ref{fig:result_simulator_parallel} shows the searching progress of our method under different number of parallel queries.
When the number of parallel queries is only one, i.e., single query per iteration, the average discrepancy curve is similar to that of the GP-based method (see Fig.~\ref{fig:result_simulator_convergence}).
As the number of parallel queries increases, our method can achieve lower discrepancy, which suggests the necessity of parallel queries for Thompson sampling to train BNNs.

\textbf{Discrepancy Reduction.}
Fig.~\ref{fig:result_online_evaluation_traffic} shows the sim-to-real discrepancy achieved by our method under different user traffic.
Here, we emulate a maximum of four mobile users in the slice, due to the limitation of hardware capability and software stability, e.g., constrained uplink throughput, in the system prototype.
Note that, the optimal simulation parameters are derived only based on the user traffic 1, and apply to the learning-based simulator for all user traffic simulations.
With the simulation parameters in Table~\ref{tb:offline_augmented_simulator}, the sim-to-real discrepancy is reduced to 81.2\%, 56.7\%, 43.6\%, and 61.6\% under the user traffic 1, 2, 3, and 4, respectively. 
This result suggests that the sim-to-real discrepancy may share some common patterns under different scenarios.
Fig.~\ref{fig:result_motivation_sim_to_real_mar_aug_gain_resource_all} shows that the sim-to-real discrepancy is reduced substantially (79.3\% in average) for almost all the resources.
Another interpretation of these figures is that, the sim-to-real discrepancy is not identical and not even under different scenarios.
As it is impractical to have online collections under all possible scenarios, the sim-to-real discrepancy will still exist between the augmented simulator and real networks.

\begin{figure*}[!t] 
\captionsetup{justification=centering}
  \begin{minipage}[t]{0.245\textwidth}
    \centering
    \includegraphics[width=1.75in, height=1.2in]{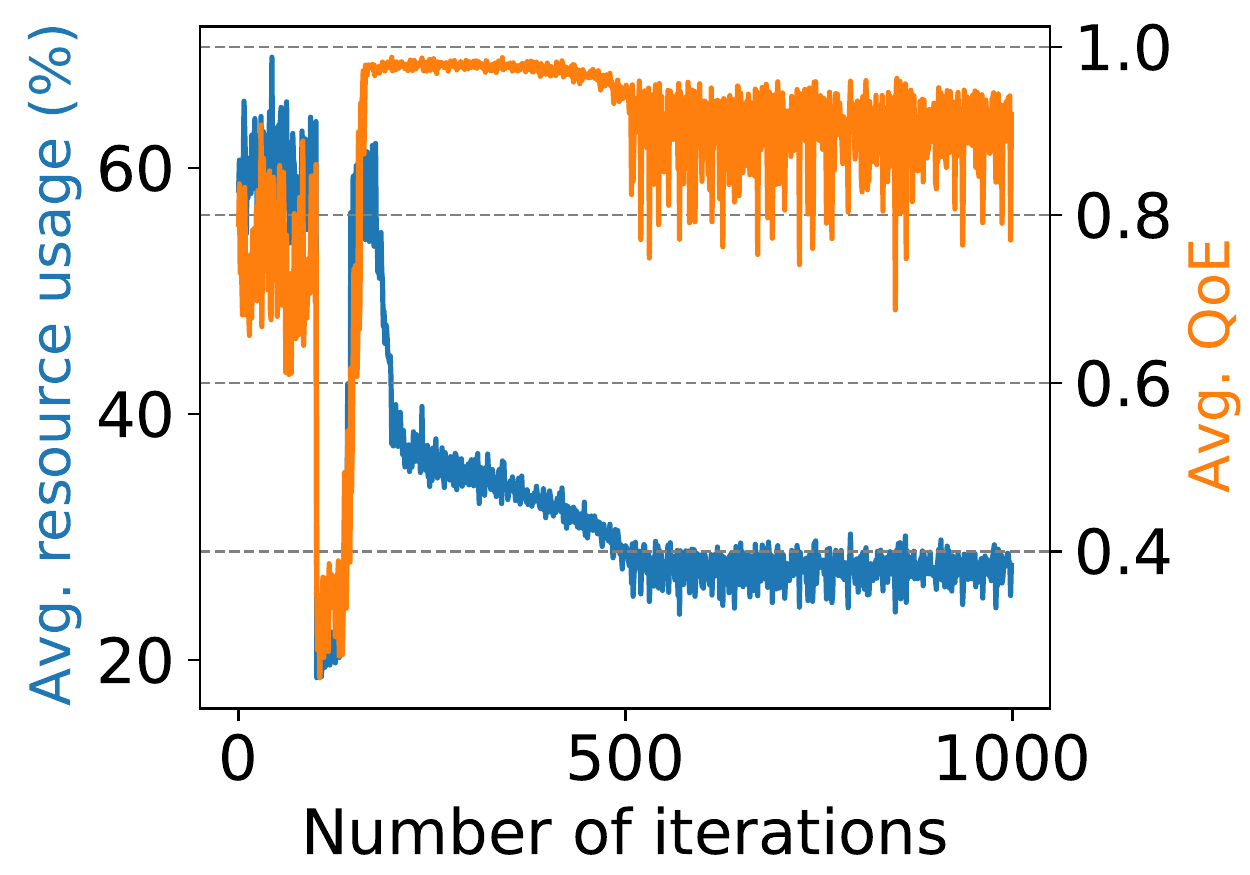}
    \caption{\small Training progress of our method}
    \label{fig:result_offline_training_convergence}
  \end{minipage}
  \begin{minipage}[t]{0.245\textwidth}
    \centering
    \includegraphics[width=1.75in, height=1.2in]{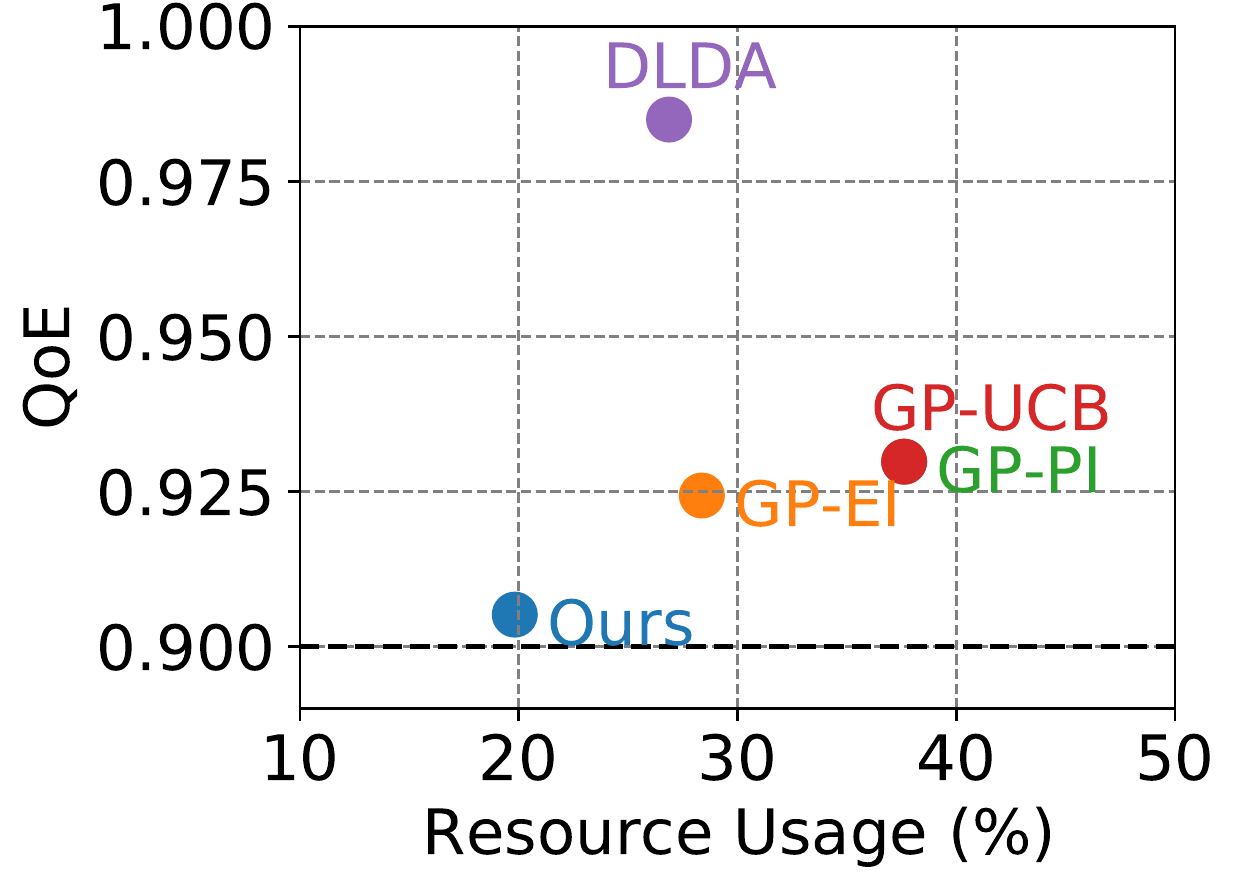}
    \captionof{figure}{\small Performance of different methods}
    \label{fig:result_offline_training_pareto}
  \end{minipage}
  \begin{minipage}[t]{0.245\textwidth}
    \centering
    \includegraphics[width=1.75in, height=1.2in]{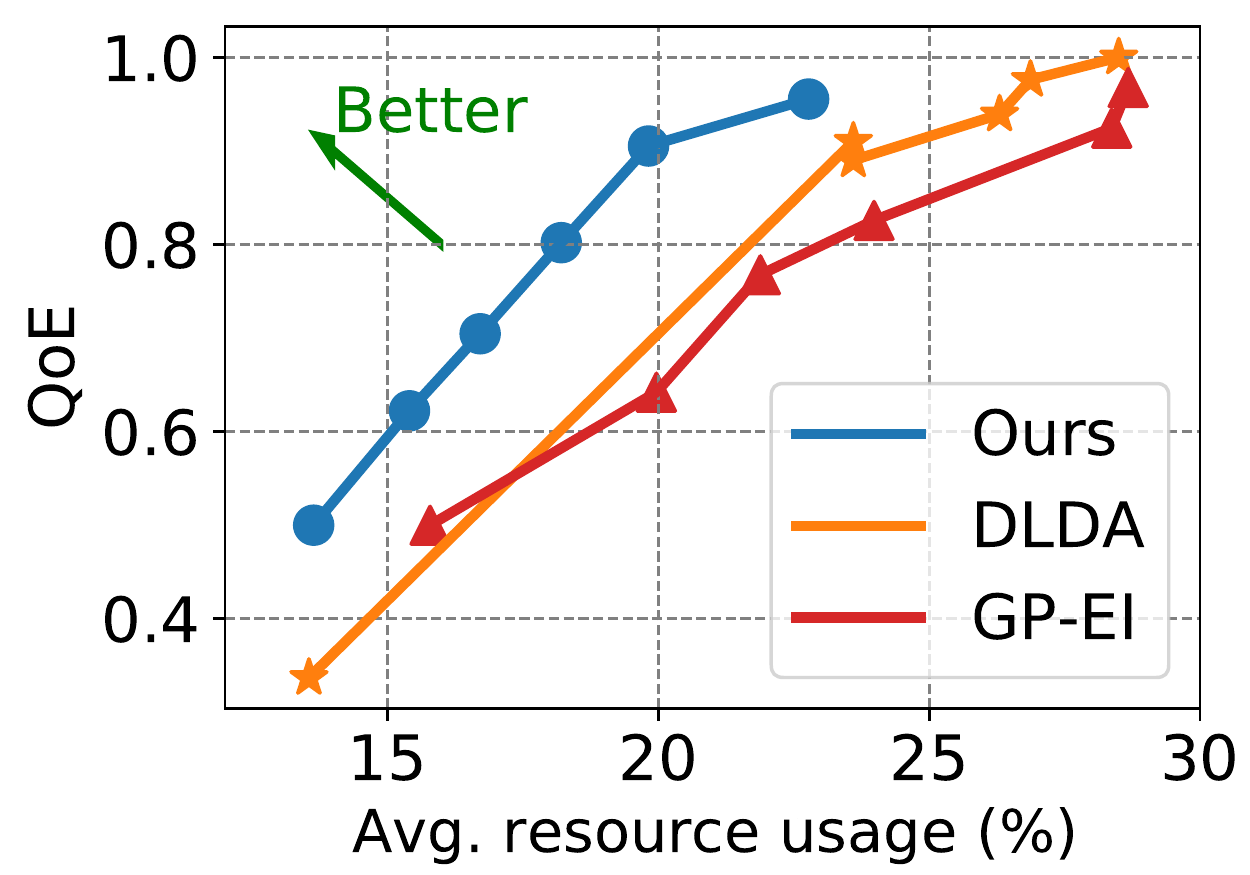}
    \captionof{figure}{\small Pareto boundary under different availability}
    \label{fig:result_offline_training_availability}
  \end{minipage}
  \begin{minipage}[t]{0.245\textwidth}
    \centering
    \includegraphics[width=1.75in, height=1.2in]{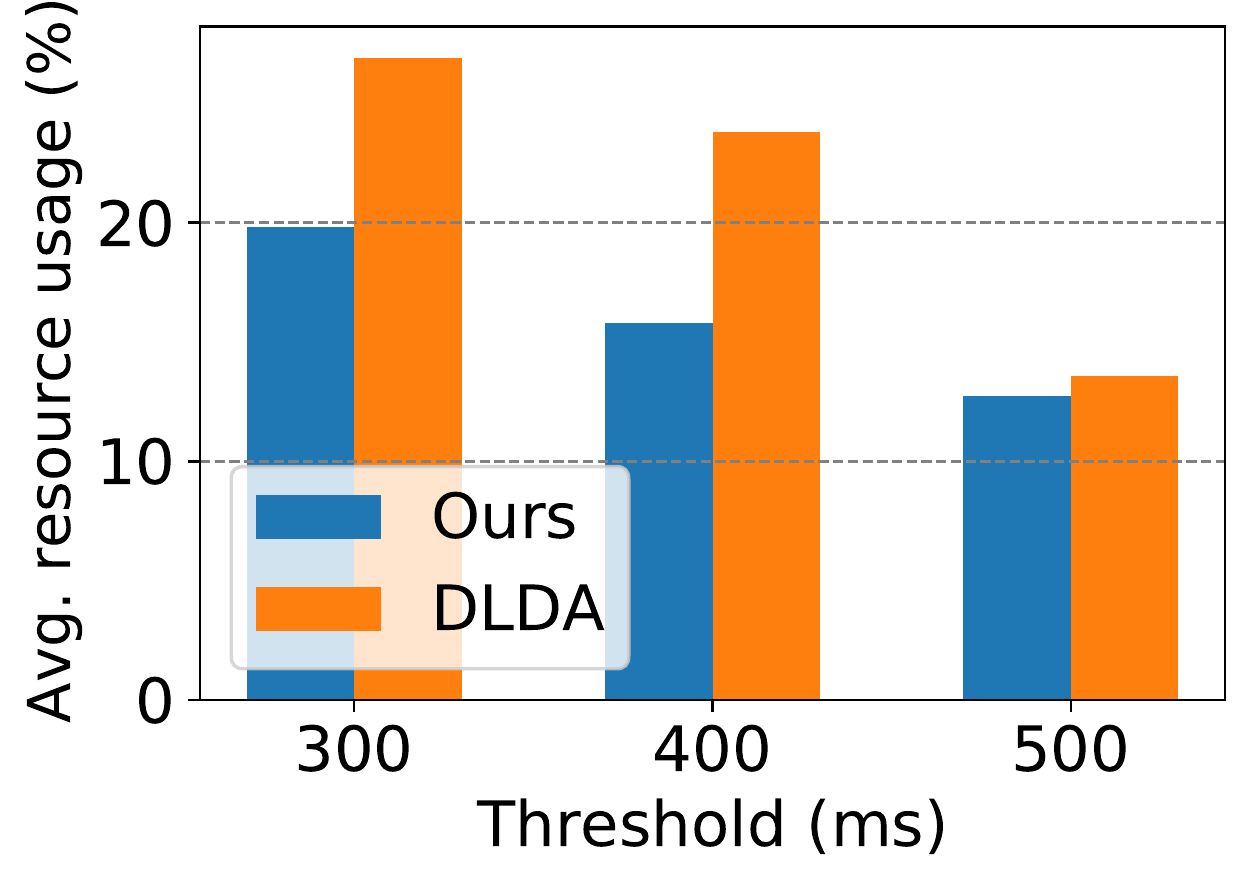}
    \caption{\small Average usage under different thresholds}
    \label{fig:result_offline_training_threshold}
  \end{minipage}
\end{figure*}

\vspace{-0.1in} \subsection{Offline Training}
\vspace{-0.02in}
In the offline training stage, we evaluate the achieved resource usage and slice QoE of the network configuration policy under given number of iterations.

\textbf{Training Progress.}
Fig.~\ref{fig:result_offline_training_convergence} shows the training progress of our method, including the average resource usage and QoE, where the first 100 iterations are purely exploration.
We observe that resource usage is gradually decreased when the average QoE is above the requirement ($E=0.9$), e.g., between the iteration 300 to 500.
Then, both the average QoE and resource usage converge, where the found optimal policy achieves stable average performance and the oscillations are mainly from Thompson sampling.
The total real-world time consumption for 1000 iterations is 7.72 hours, given the average of 27.8 seconds for NS-3 simulations.

\textbf{Performance Comparison.}
Fig.~\ref{fig:result_offline_training_pareto} depicts the QoE and resource usage under the best network configuration policy obtained by different methods.
We see that our method achieves the best performance, where the QoE is 0.905 and the resource usage is 19.81\%.
When user traffic is 1, the best configuration actions are 9 and 3 uplink and downlink PRBs, 6.2 Mbps backhaul bandwidth and 0.8 CPU ratio at the edge server, where the uplink and downlink MCS offsets are zeros.
In contrast, DLDA achieves 0.98 QoE at the cost of 26.87\% resource usage, which fails to balance the QoE and the resource usage.
The other GP-based methods, e.g., GP-EI and GP-PI, maintain more than 0.92 QoE while using up to 37.62\% resources.

\textbf{Pareto Boundary.}
Fig.~\ref{fig:result_offline_training_availability} shows the Pareto boundary of different methods, which are obtained by varying the QoE requirement $E$.
It can be seen that our method outperforms the other methods, in both QoE requirement and resource usage, where the slice SLAs on these curves are met.
For training the DLDA, we collect the offline dataset with 4096 configuration actions\footnote{We collect the slice performances in 60 seconds for each configuration action, which consumes approximately 68.5 hours in total.} by grid searching the configuration space (Eq.~\ref{prob1:const2}), where each dimension is with 4 different values (i.e., [0.0, 0.3, 0.6, 0.9]).
We notice that DLDA has a huge leap of QoE from 0.33 to 0.89, which may be attributed to the coarse-grained dataset and implies that the grid searching method fails in handling high-dimensional action spaces.

\textbf{Slice Requirement.}
Fig.~\ref{fig:result_offline_training_threshold} shows the achieved performance by our method and DLDA under different latency thresholds $Y$.
Our method obtains lower resource usage under all the scenarios with satisfied thresholds.
Besides, we see that the difference in resource usage between our method and DLDA shrinks when the threshold increases.
This is because we set a minimum of 6 uplink and 3 downlink PRBs for maintaining radio connectivities of users, where these resources may be sufficient to satisfy the loosen thresholds.

\vspace{-0.1in} \subsection{Online Learning}
In the online learning stage, we evaluate the regret of resource usage and slice QoE under given number of iterations.

\begin{figure*}[!t] 
\captionsetup{justification=centering}
  \begin{minipage}[t]{0.245\textwidth}
    \centering
    \includegraphics[width=1.75in, height=1.1in]{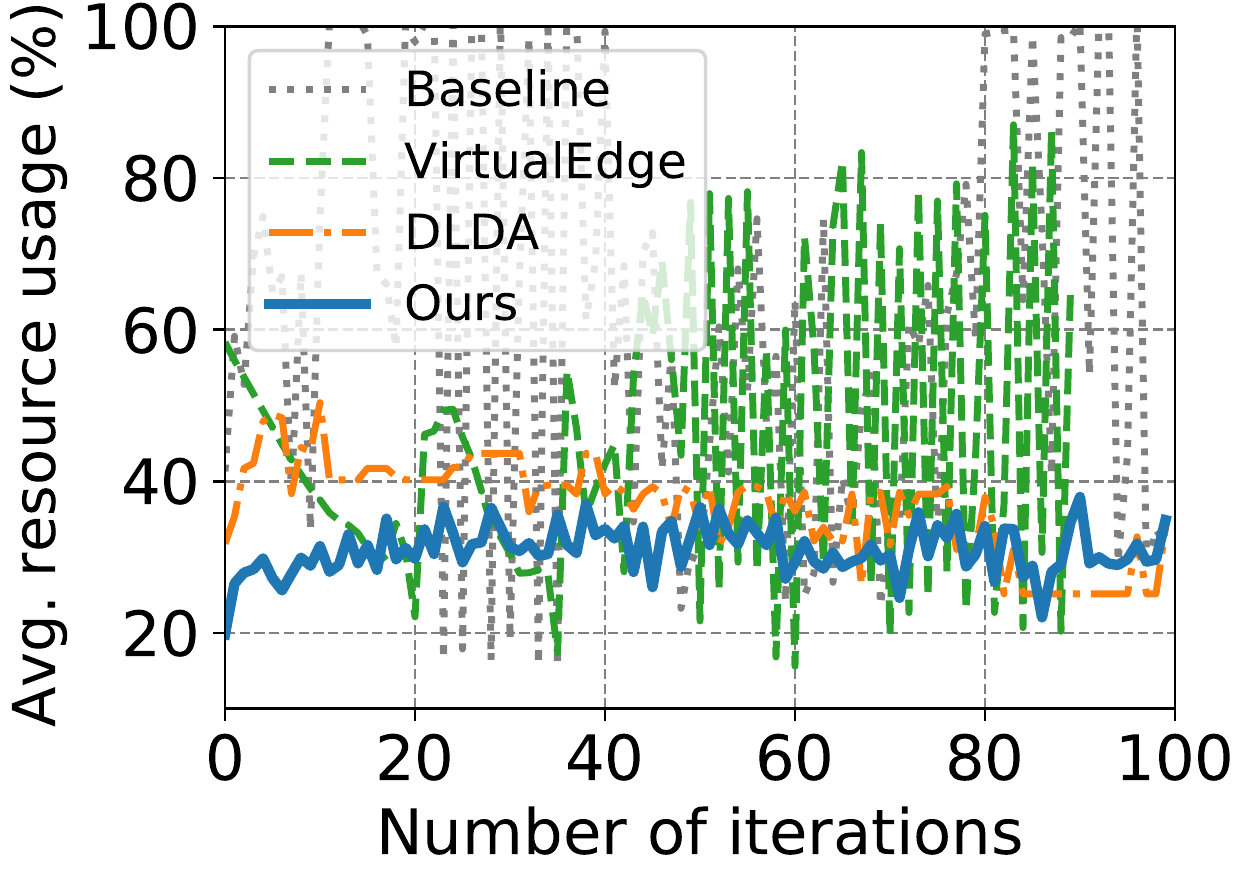}
    \caption{\small Training progress of proposed method }
    \label{fig:result_online_convergence_usage}
  \end{minipage}
  \begin{minipage}[t]{0.245\textwidth}
    \centering
    \includegraphics[width=1.75in, height=1.1in]{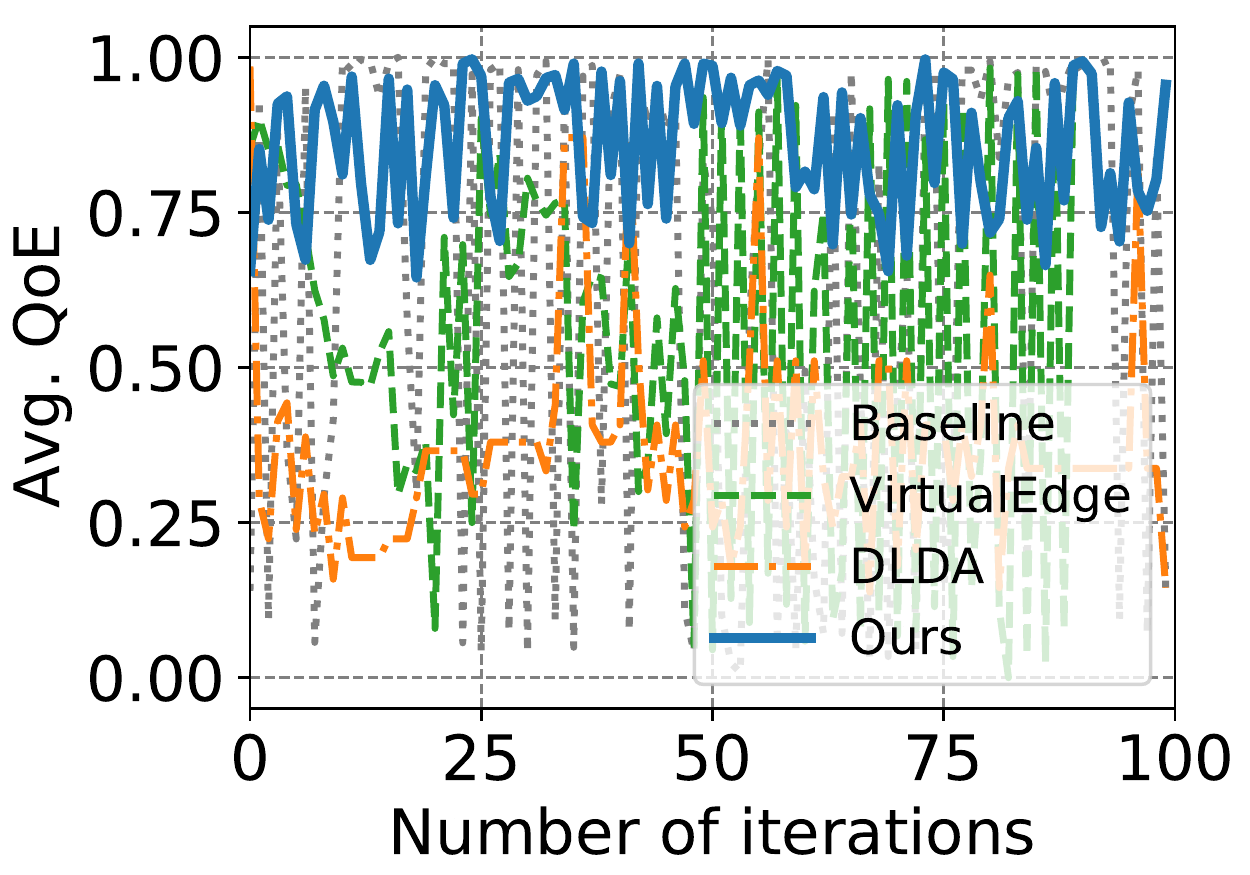}
    \captionof{figure}{\small Training progress of different methods}
    \label{fig:result_online_convergence_qoe}
  \end{minipage}
  \begin{minipage}[t]{0.245\textwidth}
    \centering
    \includegraphics[width=1.75in, height=1.1in]{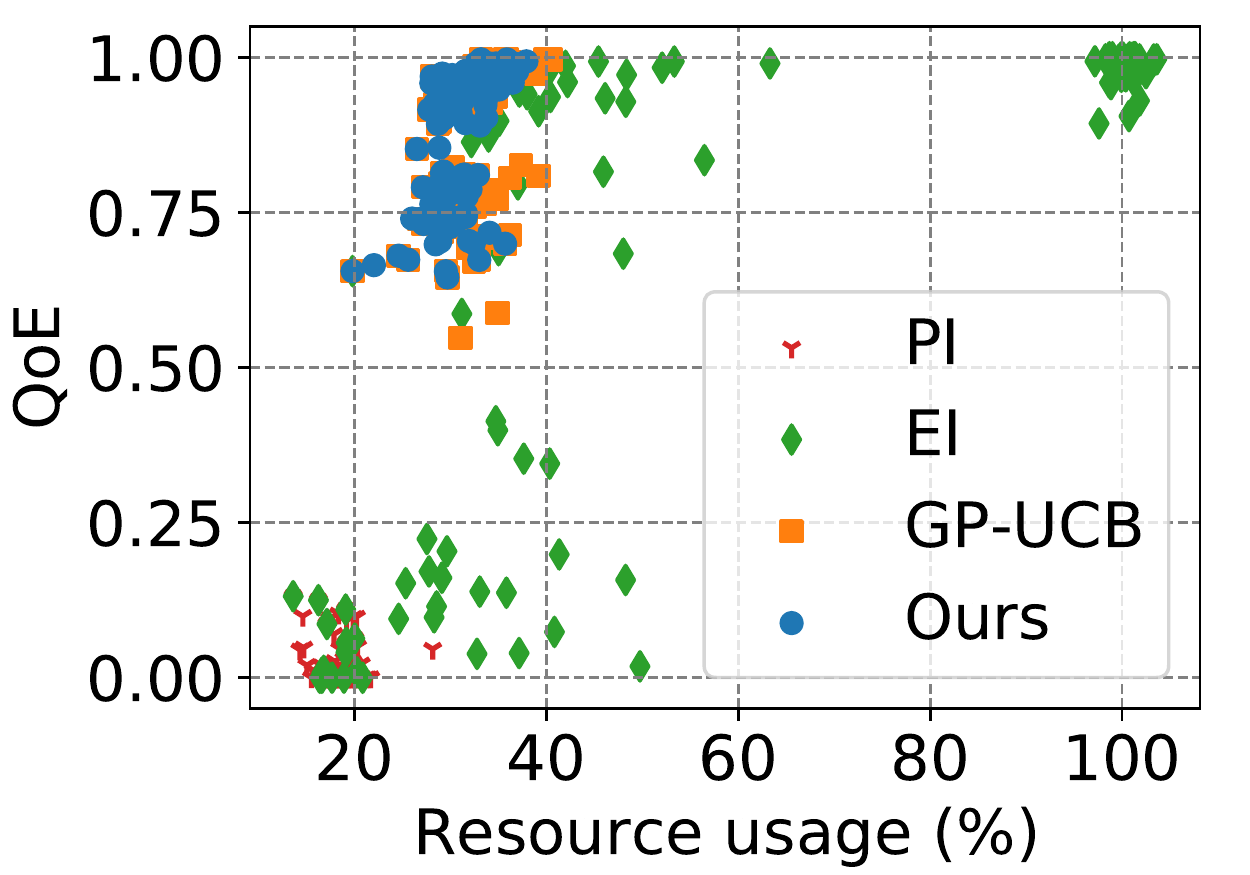}
    \captionof{figure}{\small Our method under acquisition functions}
    \label{fig:result_online_acq_function}
  \end{minipage}
  \begin{minipage}[t]{0.245\textwidth}
    \centering
    \includegraphics[width=1.75in, height=1.1in]{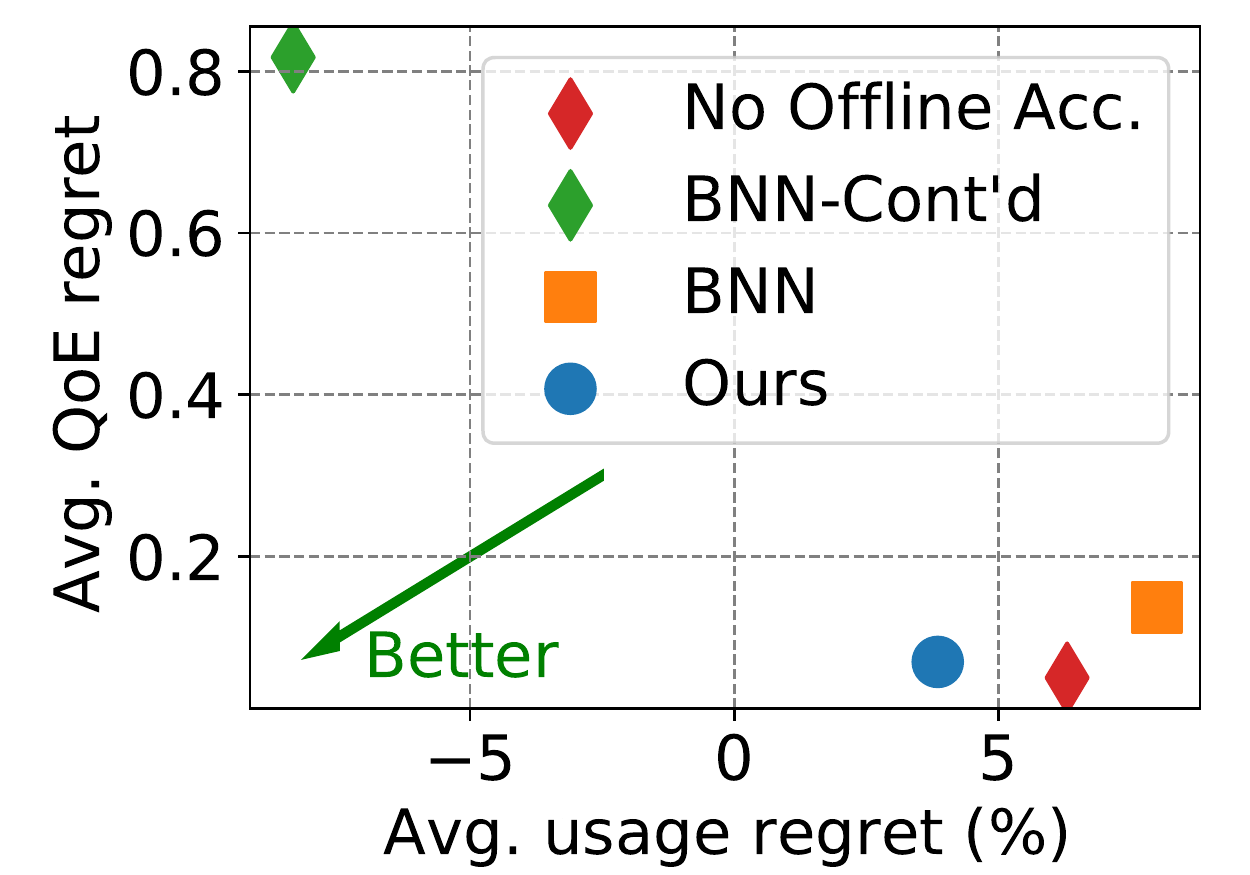}
    \caption{\small Our method under different online models}
    \label{fig:result_online_gap_model}
  \end{minipage}
\end{figure*}

\textbf{Training Progress.}
Fig.~\ref{fig:result_online_convergence_usage} and Fig.~\ref{fig:result_online_convergence_qoe} show average resource usage and slice QoE achieved during 100 online interactions, respectively.
The very first online configuration action is the optimal one obtained in the offline learning stage, if applicable, for all methods.
Recall that our method achieved 0.905 QoE in the augmented simulator (see Fig.~\ref{fig:result_offline_training_pareto}), it turns out to be 0.65 QoE in real networks, which implies the noticeable sim-to-real discrepancy. 

We see our method searches for the optimal online policy under 40\% resource usage, while the resulted slice QoEs are around the requirement ($E=0.9$).
As compared to other methods in Table~\ref{tb:online_learning_regret}, our method achieves the lowest regret for both average resource usage (63.9\% reduction than DLDA) and slice QoE (85.7\% reduction than DLDA) in 100 iterations.
In other words, our method uses only 3.16\% more resource usage and obtains 0.077 less slice QoE on average, as compared to the best policy in 100 iterations.
This result validates the efficacy of our method in terms of policy safety and sample efficiency.
Although our method needs $N=20$ offline queries with the augmented simulator after each online action, it does not affect and delay the behavior of online network configuration, especially when configuration intervals are usually tens of minutes or even hours in real networks~\cite{marquez2018should, shi2021adapting}.

\begin{table}[!t]
\small
    \begin{tabular}[b]{ m{1.8cm} | m{1.5cm}| m{1.5cm} |m{1.5cm} }\hline
       \textbf{Methods}                     &  \textbf{Avg. usage regret (\%)} &  \textbf{Avg. QoE regret}  &  \textbf{Offline queries} \\ \hline
       \textbf{ Baseline}                   & 35.83  & 0.31 &      0     \\ 
       \textbf{ VirtualEdge}                & 16.06  & 0.34 &  0   \\ 
       \textbf{ DLDA}                       & 8.79   & 0.54  &     0   \\      
       \textbf{ Ours}                       & \textbf{3.17}   & \textbf{0.077}  &     20$\times$100   \\  \hline    
    \end{tabular}
    \captionof{table}{\small Details of online learning under different methods}
\label{tb:online_learning_regret}
\end{table}

\textbf{Acquisition Function.}
Fig.~\ref{fig:result_online_acq_function} shows the footprint of our method under different acquisition functions, which is obtained by scattering the achieved resource usage and slice QoE in each iteration.
Our proposed conservative acquisition function (Eq.~\ref{eq:conservative_acq}) outperforms classic acquisition functions, which explores configuration actions with lower resource usages while approximating the slice QoE requirement.
Also, we see that GP-UCB has nearly comparative performance, which usually uses more network resources to satisfy the slice QoE requirement.
As the only difference is acquisition function, this result justifies the safe exploration of Atlas.

\textbf{Approximation Function.}
Fig.~\ref{fig:result_online_gap_model} shows the performance of our method under different approximation functions, e.g., BNN and GP.
When using BNN to approximate the sim-to-real performance difference ($G(\psi)$ in Eq.~\ref{eq:perf_real_equal_sim_plus_gap}), we see the regret of resource usage and slice QoE is increased by 107.6\% and 96.5\%, respectively.
This can be attributed to the poor sample efficiency of BNN, as 100 online collections are insufficient to train the BNN.
Alternatively, if we use the offline trained BNN to continue learning in the online learning stage (i.e., BNN-Cont'd), we see the average QoE regret soared.
Besides, if we do not exploit the offline acceleration, we observe average usage regret is increased by 63.5\% consequently.

\begin{figure*}[!t] 
\captionsetup{justification=centering}
  \begin{minipage}[t]{0.245\textwidth}
    \centering
    \includegraphics[width=1.75in, height=1.1in]{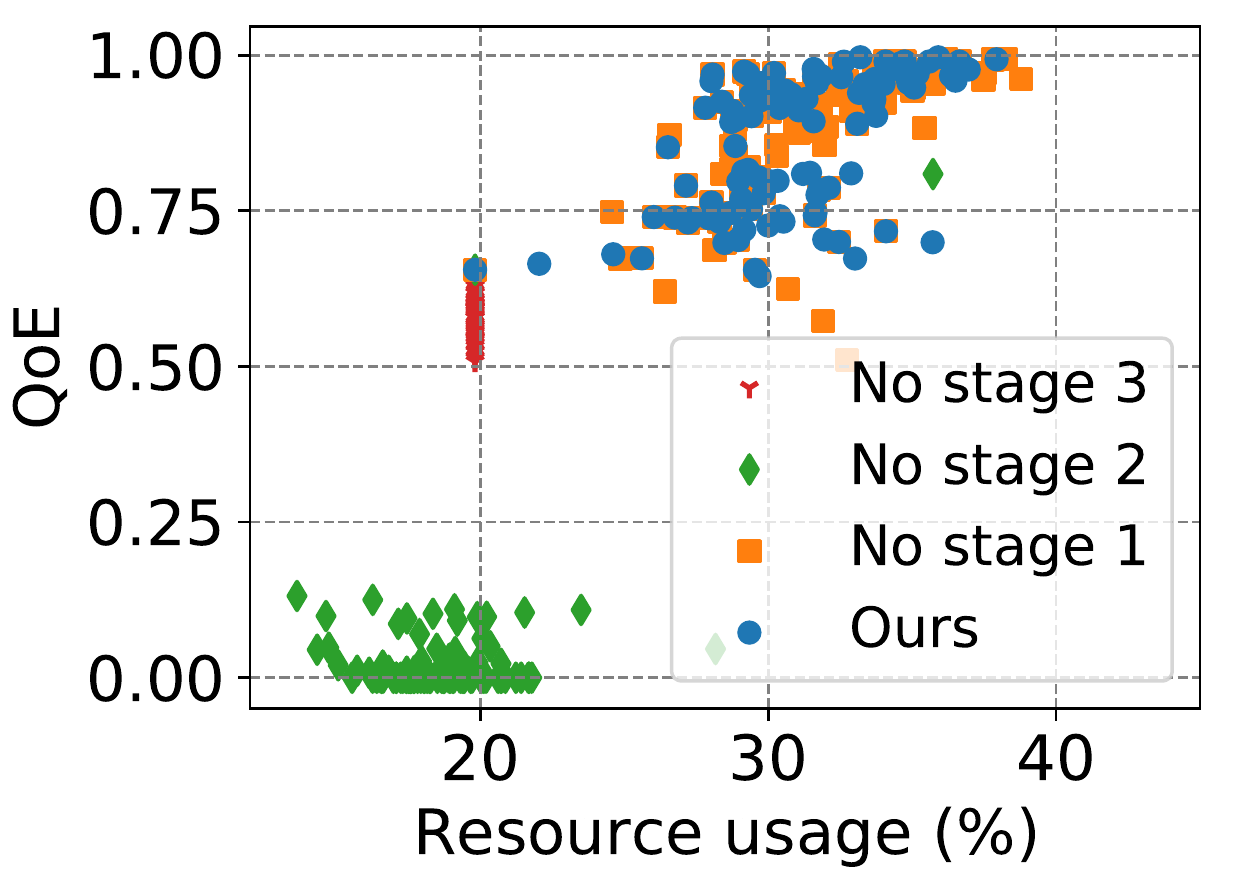}
    \caption{\small Impact of individual components}
    \label{fig:result_online_components}
  \end{minipage}
  \begin{minipage}[t]{0.245\textwidth}
    \centering
    \includegraphics[width=1.75in, height=1.1in]{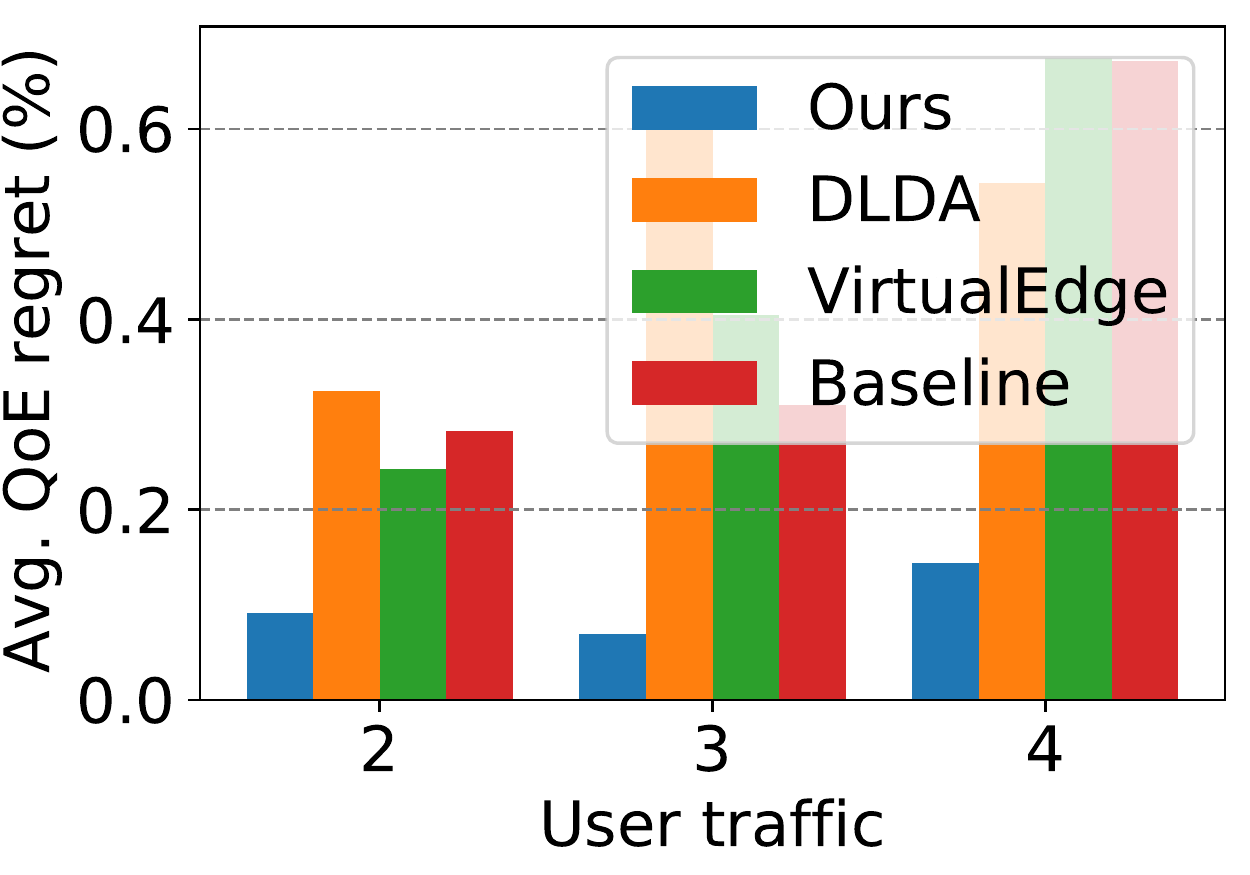}
    \captionof{figure}{\small Avg. QoE regret under different user traffic}
    \label{fig:result_online_learning_traffic_qoe}
  \end{minipage}
  \begin{minipage}[t]{0.245\textwidth}
    \centering
    \includegraphics[width=1.75in, height=1.1in]{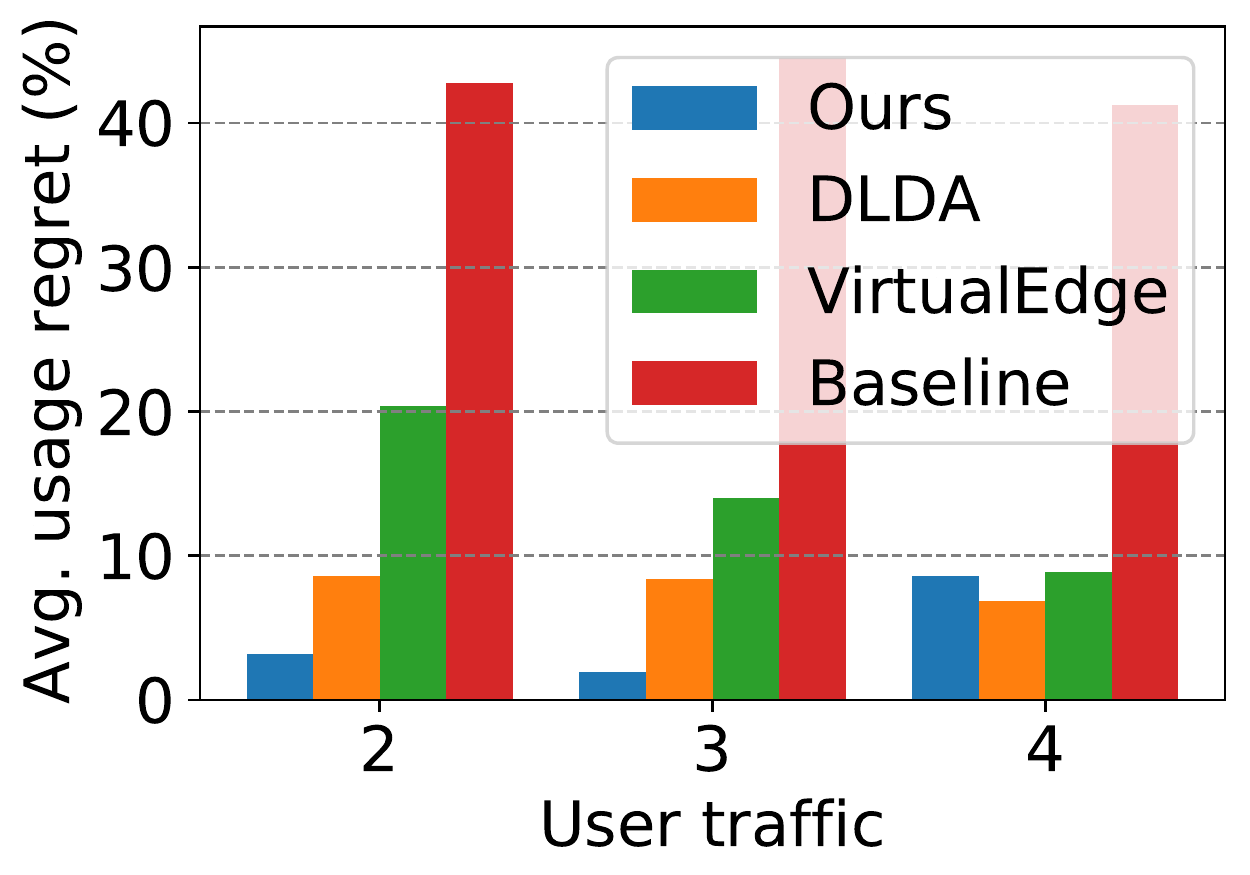}
    \captionof{figure}{\small Avg. usage regret under different user traffic}
    \label{fig:result_online_learning_traffic_usage}
  \end{minipage}
\end{figure*}

\textbf{Individual Components.}
Fig.~\ref{fig:result_online_components} shows the footprint of our method when individual stages are absent.
When there is no online learning stage, we see the resource usage keep constant while the QoE is around 0.65, where the sim-to-real discrepancy remains.
When there is no offline training stage, all the policies have to be learned via directly interacting with real networks.
As a result, it is no surprise that the early-stage performance is poor, which justifies the importance of offline training.
Without the learning-based simulator, the QoE performance is worsen due to the large sim-to-real discrepancy between original simulator and real networks.

\textbf{Dynamic Traffic.}
Fig.~\ref{fig:result_online_learning_traffic_qoe} and Fig.~\ref{fig:result_online_learning_traffic_usage} show the average slice QoE and resource usage under different user traffic, where the latency threshold $Y=500ms$.
We see our method achieves the lowest regret of both resource usage and slice QoE for almost all user traffic.
Although DLDA has a slightly lower regret of average resource usage when user traffic is 4, its average QoE regret is much higher than that of ours.


\vspace{-0.17in} \section{Related Work}
\vspace{-0.02in}

\textbf{Network Slicing.}
Network slicing systems have been increasingly studied to support heterogeneous applications in terms of flexibility and cost-efficiency~\cite{foukas2016flexran, bega2019machine, liu2019virtualedge, salvat2018overbooking, d2020sl}.
Foukas \emph{et. al.}~\cite{foukas2017orion} designed Orion as the first RAN slicing system based on FlexRAN~\cite{foukas2016flexran}, which enables on-the-fly RAN virtualization with both performance and functional isolation.
Marqueze \emph{et. al.}~\cite{marquez2018should} empirically demonstrated that dynamic resource orchestration improves the efficiency of resource multiplexing in RAN.
Liu \emph{et. al.}~\cite{liu2019virtualedge} proposed VirtualEdge that automatically learns to orchestrate cross-domain resources for maximizing the performance of slices.
Salvat \emph{et. al.}~\cite{salvat2018overbooking} proposed to multiplex network resources via overbooking for concurrently supporting multiple slices with two new resource provisioning algorithms.
D'Oro \emph{et. al.}~\cite{d2020sl} proposed Sl-EDGE that enables joint network-edge slicing, where near-optimal algorithms are designed to instantiate slices under constrained resources in edge nodes. 
These works, however, mainly focused on network slicing in individual domains with known slice resource demands, which do not apply to network configuration in end-to-end slicing systems.

\textbf{Machine learning for Networking:}
Recent advances in machine learning techniques provoke the interest in applying ML in handling complex networking problems~\cite{lee2020perceive, liu2021constraint, yan2020learning, wang2020job, aggarwal2020libra}.
Microscope~\cite{zhang2020microscope} introduced an ML-based decomposition method to efficiently estimate the service demand of slices, which deals with complex spatiotemporal features in aggregated traffic.
EdgeSlice~\cite{liu2020edgeslice} handled the multi-user and multi-domain resource orchestration problems by using a decentralized deep reinforcement learning (DRL) algorithm for network slicing.
The constrained resource allocation problems are studied in network slicing~\cite{khairy2020constrained, liu2020edgeslice}, where different techniques, e.g., reward shaping and interior-point method~\cite{liu2020constrained}, are utilized to maintain the performance requirement of slices.
These works, however, primarily focused on the offline training and online deployment strategy, which suffer the sim-to-real discrepancy when applying offline policies to real networks.

\textbf{Sim-to-Real Discrepancy.}
The sim-to-real discrepancy has been increasingly unveiled for online system design.
Mao \emph{et. al.}~\cite{mao2019learning} identified the non-negligible performance difference between the high-fidelity simulator and real data center networks under the identical policy.
OnRL~\cite{zhang2020onrl} is proposed to address the sim-to-real discrepancy in real-time mobile video telephony, via online DRL within real networks, where an individualized hybrid learning method is designed to counter the potential performance degradation.
OnSlicing~\cite{liu2021onslicing} resolved the cross-domain resource orchestration problem in online network slicing by designing a proactive baseline switching mechanism for near-zero SLA violations.
These works focus on time-correlated problems with small time scales (e.g., seconds or minutes) and unfortunately fail in network configuration problems with large time scales. 
Shi \emph{et. al.}~\cite{shi2021adapting} designed a transfer learning DNN-based algorithm to bridge the sim-to-real gap by using both offline and online datasets (obtained by grid searching), which fails in terms of performance assurance and sample efficiency.


\vspace{-0.17in} \section{Discussion}
\vspace{-0.02in}

\textbf{Scalability.}
As Atlas tackles the service configuration of individual network slices, it can seamlessly accommodate dynamic slice admission and removal.
When a new slice is admitted and launched, an individualized Atlas will be initialized to build its learning-based simulator, offline train the policy, and online learn to configure continually.
As there are infrastructure changes, e.g., installing more antennas and higher bandwidth switches, the corresponding parts in the learn-based simulator will be updated.
The simulation parameters will be continually searched based on its last optima if needed. 
The offline policies will be fine-tuned in the updated simulator, and the online learning stage continues uninterruptedly as the GP model learns only the sim-to-real discrepancy. 
All these procedures will be offline executed and accelerated if applicable, without the need for additional online transitions.

\textbf{Adaptability.} 
For parameter and configuration space changes, Atlas can reuse previous experience (e.g., buffered transitions) to accelerate the training and finetuning of new approximation functions.
If network dynamics are observable, e.g., spatiotemporal traffic changes, the topology and settings in the simulator will be updated accordingly.
Then, the offline simulator and training stage in Atlas will be refined, e.g., simulation parameters and approximation functions.
If network dynamics are not observable, e.g., instant link failure, Atlas will recognize it as the fluctuations of simulation-to-reality discrepancy and rely on the sample efficient Gaussian process model to learn and adapt continually.

\textbf{Generalizability.} 
The proposed methods in Atlas have the potential to be generalized to tackle sim-to-real discrepancy and online learning problems in other network systems, e.g., datacenter networks and wireless mesh networks~\cite{shi2021adapting}.
This is attributed to the minimal assumptions of Atlas, summarized as follows:
1) there is a queryable simulated environment to imitate the real-world system.
2) there is an online dataset collected directly from the real-world system.
3) the real-world system can be queried, where network states and generated results are observable.

\vspace{-0.12in} \section{Conclusion}
\vspace{-0.02in}
In this paper, we proposed an online network slicing system to automate the service configuration of slices, with three novel interrelated stages.
First, we designed a learning-based simulator to reduce the sim-to-real discrepancy with a new parameter searching method.
Second, we designed a novel offline training algorithm to train the policy in the augmented simulator.
Third, we designed a novel online learning algorithm to safely learn the policy and resolve the sim-to-real discrepancy with real networks.
We develop an end-to-end network prototype and evaluated Atlas in terms of performance assurance, sample efficiency, and resource usage. 

\vspace{-0.12in}
\begin{acks}
\vspace{-0.02in}
This work is partially supported by the US National Science
Foundation under Grant No. 2212050, No. 2147624, No. 2147821, and No. 2147623.
\end{acks}

\bibliographystyle{acm}
\bibliography{ref/reference.bib, ref/qiang.bib}

\newpage

\appendix

\vspace{-0.1in} \section{Parameter Searching Method}
\label{sec:appendix:simulator}
The parameter searching algorithm for learning-based simulator is summarized in Alg.~\ref{alg:offline_sim}.
First, tens of thousands of simulation parameters are sampled from the parameter space (i.e., Eq.~\ref{prob0:const1}).
Second, the next query is selected by minimizing the weighted discrepancy, where the sim-to-real discrepancy ${KL}[\mathcal{D}_r || \mathcal{D}_s (\mathbf{x})]$ is estimated by the BNN for all the sampled parameters.
Third, the actual values of sim-to-real discrepancy is obtained via querying the offline simulation.
Fourth, the transition $<\mathbf{x}, {KL}[\mathcal{D}_r || \mathcal{D}_s (\mathbf{x})]>$ are stored for BNN training.
Note that these above steps are executed in parallel with multiprocessing technique.

\begin{algorithm}[!h]
	\caption{The Parameter Searching Method}\label{alg:offline_sim}

	\KwIn{$H$, $\alpha$, $\mathbf{\hat{x}}$ }

	\While{True}
	{
        \For{$n=0,1,... (Parallel)$ }
        {
            Sample $\mathbf{x}$\;
            Estimated sim-to-real discrepancy ${KL}[\mathcal{D}_r || \mathcal{D}_s (\mathbf{x})]$ from BNN for all samples\;
            Get the next query $\mathbf{x} \gets \arg\min \left({{KL}[\mathcal{D}_r || \mathcal{D}_s (\mathbf{x})]  +  \alpha |\mathbf{x} - \mathbf{\hat{x}}|_2} \right)$\;
            Obtain sim-to-real discrepancy ${KL}[\mathcal{D}_r || \mathcal{D}_s (\mathbf{x})]$ by querying simulator\;
            Store transition $<\mathbf{x}, {KL}[\mathcal{D}_r || \mathcal{D}_s (\mathbf{x})]>$\;
            Train the BNN with new added transitions\;
        }
        \If{ Convergence}
        {
            $\mathbf{break}$\;
        }
	}
\end{algorithm}

\vspace{-0.1in} \section{The Offline Algorithm}
\label{sec:appendix:offline}
The offline network configuration algorithm is summarized in Alg.~\ref{alg:offline_orch}.
First, tens of thousands of network configuration actions are sampled from the configuration space (i.e., Eq.~\ref{prob1:const2}).
Second, the next query is selected by minimizing the Lagrangian, where the resource usage is calculated in $F(\phi)$ and the slice QoE is estimated by the BNN for all the sampled actions.
Third, the actual values of slice QoE is obtained via querying the offline simulation.
Fourth, the transition $<\mathbf{a}_t, F(\phi), Q_s(\phi)>$ are stored for BNN training.
Note that these above steps are executed in parallel with multiprocessing technique.
Then, the multiplier is updated by averaging the results from parallel queries. 

\begin{algorithm}[!h]
	\caption{The Offline Algorithm}\label{alg:offline_orch}

	\KwIn{$\varepsilon$, $E, Y, A$,  }
	Initialize parameters $\lambda \gets 0$\;

	\While{True}
	{
        \For{$n=0,1,... (Parallel)$ }
        {
            Sample $\mathbf{a}_t$, calculate $F(\phi)$ for all samples\;
            Estimated $Q_s(\phi)$ from BNN for all samples\;
            Calculate Lagrangian in Eq.~\ref{eq:lagrangian}\;
            Get the next query $\mathbf{a}_t \gets \arg\min {L}$\;
            Obtain slice QoE by querying the simulator\;
            Store transition $<\mathbf{a}_t, F(\phi), Q_s(\phi)>$\;
            Train the BNN with new added transitions\;
        }
        Update multiplier $\lambda \gets $ Eq.~\ref{eq:update_multiplier}\;
        \If{ Convergence}
        {
            $\mathbf{break}$\;
        }
	}
	\Return{$\phi$}\;
\end{algorithm}

\vspace{-0.1in} \section{The Online Algorithm}
\label{sec:appendix:online}
The online network configuration algorithm is summarized in Alg.~\ref{alg:online_orch}.
First, the multiplier in the online learning stage is initialized with the final multiplier in the offline training stage.
Then, we update the multiplier by interacting with the offline simulator for $N$ times (Line~\ref{alg:line:start}-\ref{alg:line:end}).
Specifically, we calculate the Lagrangian for all sampled configuration actions with both the offline BNN and the online GP model.
The next network configuration is selected by minimizing the Lagrangian.
We update the multiplier with $\lambda \gets $ Eq.~\ref{eq:update_multiplier_revised} by querying offline simulator and estimating GP model.
Next, we online query real networks, when it needs to, e.g., configuration interval is 1 hour.
The new obtained online transitions are used to update the GP model for approximating the sim-to-real performance differences.

\begin{algorithm}[!h]
	\caption{The Online Algorithm}\label{alg:online_orch}

	\KwIn{$\varepsilon$, $Y, E, A $, $B$, $N$  }
	Initialize multiplier $\lambda$ from offline stage\;

	\While{True}
	{
	    \For{$n=0,1,... N$ }
	    {   
            Sample $\mathbf{a}_t$, calculate $F(\phi)$ for all samples\label{alg:line:start}\;
            Estimated $Q(\phi)$ from BNN $Q_s(\phi)$ and GP model $G(\psi)$ for all samples with Eq.~\ref{eq:perf_real_equal_sim_plus_gap}\;
            Calculate Lagrangian in Eq.~\ref{eq:lagrangian_online}\;
            Get the next query $\mathbf{a}_t \gets \arg\min {L}$\;
            Obtain $Q_s(\phi)$ by querying simulator\;
            Estimate $G(\psi)$ with $\psi$\;
            Update multiplier $\lambda \gets $ Eq.~\ref{eq:update_multiplier_revised} \label{alg:line:end}\;
        }
        \If{ Time to online query}
        {
            Apply $\mathbf{a}_t$ to real networks\;
            Calculate $G(\psi) = Q(\phi) - Q_s(\phi)$\;
            Train $\psi$ with new online transitions\;
        }
        \If{ Convergence}
        {
            $\mathbf{break}$\;
        }
	}
	\Return{$\psi$}\;
\end{algorithm}

\section{Artifact Appendix}

\subsection{Abstract}
This section describes the instructions for performing artifact evaluation for this paper.
In the paper, we achieve \emph{Atlas} with three integrated stages, i.e., learning-based simulator, offline training, and online learning. 


\subsection{Artifact check-list}

{\small
\begin{itemize}
  \item {\textbf{Algorithm:} Algorithm 1, 2, 3 described in the paper}
  \item {\textbf{Program: }Python scripts, NS-3 simulator, OpenDayLight, executable files in OpenAirInterface RAN and CORE, Android applications}
  \item {\textbf{Compilation:} g++ 8.0, gcc 8.0 or higher}
  \item {\textbf{Data set:} own dataset provided in open-source codes.}
  \item {\textbf{Run-time environment:} At least two desktop computers, one for RAN and one for CN\&Edge. OS can be either Ubuntu 18.04 or 20.04. Docker containers is needed in CN desktop.}
  \item {\textbf{Hardware:} For network simulator: No restriction on hardware; For network prototype: Intel i7 or above CPU, NI USRP B210, Ruckus ICX 7150-C12P, OnePlus 9 5G smartphone, Antenna compatible with LTE B7 frequency band.}
  \item {\textbf{Execution:} sole user is preferred}
  \item {\textbf{Metrics:} runtime log in NS-3, performance reported by smartphones.}
  \item {\textbf{Output:} saved experiment results in pickle format, which will be used for following figure plotting.}
  \item {\textbf{Experiments:} see README in the open-source codes.}
  \item {\textbf{How much disk space required (approximately)?:} ~1GB for data and trained model, the disk space for software installation are not counted.}
  \item {\textbf{How much time is needed to prepare workflow (approximately)?:} For network simulator: less than 30 minutes; For network prototype: several hours if not more.}
  \item {\textbf{How much time is needed to complete experiments (approximately)?:} For stage 1: 3~5 hours, under 16 threads; For stage 2: 3~5 hours, under 16 threads; For stage 3: 1~3 hours.}
  \item {\textbf{Publicly available?:} Yes.}
\end{itemize}
}

\subsection{Descriptions}

{\small
\begin{itemize}
  \item {$main\_simulator.py$ is the main file for conducting experiments for the stage 1: learning-based simulator.}
  \item {$main\_offline.py$ is the main file for conducting experiments for the stage 2: offline training.}
  \item {$main\_online.py$ is the main file for conducting experiments for the stage 3: online learning.}
  \item {$plot\_*.py$ are mainly for plotting the results based on completed experiments.}
  \item {$system.py$ is the main file for connecting with network prototype.}
  \item {$simulator.py$ is the main file for connecting with network simulator.}
\end{itemize}
}

\subsubsection{How to access}
The codes are open-sourced available in \\
\href{https://doi.org/10.5281/zenodo.7262492}{https://doi.org/10.5281/zenodo.7262492}, where more detailed instructions can be found in \href{https://github.com/int-unl/Atlas.git}{https://github.com/int-unl/Atlas.git}.
\subsubsection{Hardware dependencies}
At least two desktop computers with Intel i7 or above CPUs, one for RAN and one for CN\&Edge. OS can be either Ubuntu 18.04 or 20.04. Docker containers is needed in CN desktop.
\subsubsection{Software dependencies}
For executing the codes: Ubuntu 20.04, Python 3.6.9, PyTorch 1.10.2, scipy 1.5.4, sklearn 0.24.2, numpy 1.19.5, pickle 4.0, CUDA is not required.

For RAN host in the network prototype: Ubuntu 18.04, low-latency kernel; For CN and Edge in the network prototype: Ubuntu 20.04.
\subsubsection{Data sets}
We collect our own dataset from the real-world network prototype. 
The file is in $app\_eval/$ folder, which is mainly used for the stage 1: learning-based simulator. 

\subsection{Network Simulator and Prototype Build}

{\small
\begin{itemize}
    \item[] {\bf Network Simulator:}
    \item {Install NS-3 3.36, either with official instruction or provided bash file.}
    \item {Add additional files (i.e., $edge/$) in NS-3 $contrib/$ folder, and rebuild.}
    \item {Add additional files (i.e., main.cc) in NS-3 $scratch/$ folder, and rebuild.}
    \item {Validate the network simulator can be connected by $simulator.py$ script.}
    \item[] {\bf Network Prototype:}
    \item {Install OpenAirInterface RAN with official instruction.}
    \item {Install OpenAirInterface CORE with official instruction, where the dockerized network functions are required, e.g., SPGW-U and HSS.}
    \item {Modify SPGW-C and rebuild to enable core network slicing, validate it redirects specific mobile users to corresponded SPGW-Us.}
    \item {Connect smartphones with RAN with programmed USIM card to match the PLMN and other parameters, validate they can access to the RAN and the SPGW-U docker.}
    \item {Install the provided Android application to smartphones (Android 11).}
    \item {Install the provided edge server application in individual SPGW-U dockers, validate smartphones can connect with their servers with periodically performance updates.}
    \item {Install FlexRAN controller 2.0 with official instruction, validate its slicing capability when changing PRB allocation on RAN to different mobile users.}
    \item {Initialize the SDN switch, install OpenDayLight to connect the switch, validate the provided $tn\_server.py$ can connect OpenDayLight.}
    \item {Connect SDN switch between the RAN and CORE desktop, validate it can enforce bandwidth allocation to between mobile users and their SPGW-U dockers.}
    \item {As everything is well validated individually, close all scripts and programs.}
    \item {The order of running the network prototype is: OpenDayLight -- Transport controller -- FlexRAN controller -- CORE -- edge server applications -- RAN -- Configure slicing parameters -- Smartphone (disable airplane mode and open the app) -- start $system.py$.}
    \item {The demo procedures of bringing up the network prototype is online: \href{https://youtu.be/-AFw17ANBN8}{https://youtu.be/-AFw17ANBN8}}
\end{itemize}
}

\subsection{Experiment workflow}

{\small
\begin{itemize}
  \item {Download open-source codes}
  \item {Install package dependencies, e.g., sklearn, scipy, torch, and more.}
  \item[] {\bf Stage 1: learning-based simulator.}
  \item {Run $main\_simulator.py$, where the arguments vary according to different experiments.}
  \item {Run $plot\_simulator.py$ to reproduce the figures in the paper, after all corresponded experiments are done.}
  \item {Update the searched optimal simulation parameter to $parameter.py$.}
  \item[] {\bf Stage 2: offline training.}
  \item {Run $main\_offline.py$, where the arguments vary according to different experiments.}
  \item {Run $plot\_offline.py$ to reproduce the figures in the paper, after all corresponded experiments are done.}
  \item {Update the searched optimal resource configuration to $parameter.py$.}
  \item[] {\bf Stage 3: online learning.}
  \item {Bring up the network prototype (see above), validate it is live and can be connected by the $system.py$ script.}
  \item {Run $main\_online.py$, where the arguments vary according to different experiments.}
  \item {Run $plot\_online.py$ to reproduce the figures in the paper, after all corresponded experiments are done.}
\end{itemize}
}

\end{document}